\documentclass[letterpaper]{article} 
\usepackage{lineno}
\usepackage[preprint]{aaai2027}
\usepackage{graphicx} 
\usepackage{natbib}  
\usepackage{caption} 
\usepackage{algorithm}
\usepackage{algorithmic}
\usepackage{multirow}

\usepackage{newfloat}
\usepackage{listings}
\DeclareCaptionStyle{ruled}{labelfont=normalfont,labelsep=colon,strut=off} 
\floatstyle{ruled}
\newfloat{listing}{tb}{lst}{}
\floatname{listing}{Listing}

\definecolor{groupgray}{RGB}{230,230,230}
\definecolor{oursblue}{RGB}{219,238,252}

\usepackage{booktabs}
\usepackage{amsmath}
\usepackage{amssymb}
\usepackage{booktabs}
\usepackage[table]{xcolor}
\title{PsyMemBehav: Psychology-Informed and Memory-Grounded Two-Level Identity-to-Action Reasoning for Social-Media Behavior Simulation}
\author{
Wenji Zhang\textsuperscript{1,*},
Xilong Cheng\textsuperscript{1,*},
Yunxiao Qin\textsuperscript{1,2,*,\textdagger},
Mai Tu\textsuperscript{1},
Tongye Zhu\textsuperscript{1},
Yuhao Zhou\textsuperscript{1},\\
JingBo Ji\textsuperscript{1},
Yuting Tan\textsuperscript{1},
Duoyin Li\textsuperscript{1},
Hongjiang Xiao\textsuperscript{1,2,\textdagger},
Yuan Zhang\textsuperscript{1,2}
}
\affiliations{
\textsuperscript{1}Communication University of China, China\\
\textsuperscript{2}State Key Laboratory of Media Convergence and Communication, China\\
202311153001@mails.cuc.edu.cn,
chengzhengyu330@cuc.edu.cn,
qinyunxiao@cuc.edu.cn\\
\textsuperscript{*}Equal contribution.
\quad
\textsuperscript{\textdagger}Corresponding author.
}
\begin{document}

\maketitle

\begin{abstract}
Large language models (LLMs) support social simulation, but faithfully reproducing real users requires grounding both structured platform decisions and language in sparse, noisy behavioral evidence. We introduce \textbf{PsyMemBehav}, which maps psychology-informed profiles and retrieved memories to target-first platform behavior through two-level identity-to-action reasoning. To support this approach with comprehensive behavioral traces, we construct a dataset PsyMemBehav-bench from 5.57M observed behaviors of 5,862 real users (approximately 950 per user) across Reddit, Bluesky, and Twitter. Each recorded engagement is modeled as target selection followed by \textsc{Comment}, \textsc{Like}, or \textsc{Repost} and optional response generation. PsyMemBehav combines 41-dimensional user descriptors, top-5 BM25 memories, reverse-CoT supervision, and rubric-guided DPO over complete reasoning-action trajectories. Experiments show that PsyMemBehav 27B ranks first in 11 of 12 reported individual-level metrics, improving average response quality over GPT-5.4 from 54.28 to 59.68 (9.95\% relative), and obtains the lowest group-level MAE (0.081) and MSE (0.009) among simulated systems. These results support evidence-grounded user modeling for observed-thread continuation; the benchmark remains conditional on recorded engagement and does not model no-action decisions. 
\footnote{
All code, checkpoints, and data will be made publicly available immediately upon acceptance.
}
\end{abstract}

\section{Introduction}

Online public opinion develops through repeated user interactions: people comment, like, repost, choose whom to address, and express stances within evolving discussion threads. When aggregated, these micro-level decisions shape observable stance distributions, interaction networks, attention concentration, and information diffusion. Simulating this process offers a controlled setting for studying how public opinion may evolve under different discussion contexts \citep{hu2024laidsim,liu2025mosaic,jeon2025simulating,cheng2025interactive}. 
LLM agents offer a flexible mechanism for such simulations, but the credibility of aggregate results depends in part on whether individual behaviors are grounded in heterogeneous real-user evidence \citep{argyle2023out,gao2024large}. Fluent yet generic or uniformly active agents can distort action frequencies, target selection, and stance composition when their outputs are aggregated. We therefore study user-level target--action--response prediction as a behavioral foundation for evaluating thread-level public-opinion simulation in observed social-media contexts.

\begin{figure}[t]
	\centering
	\includegraphics[width=\columnwidth]{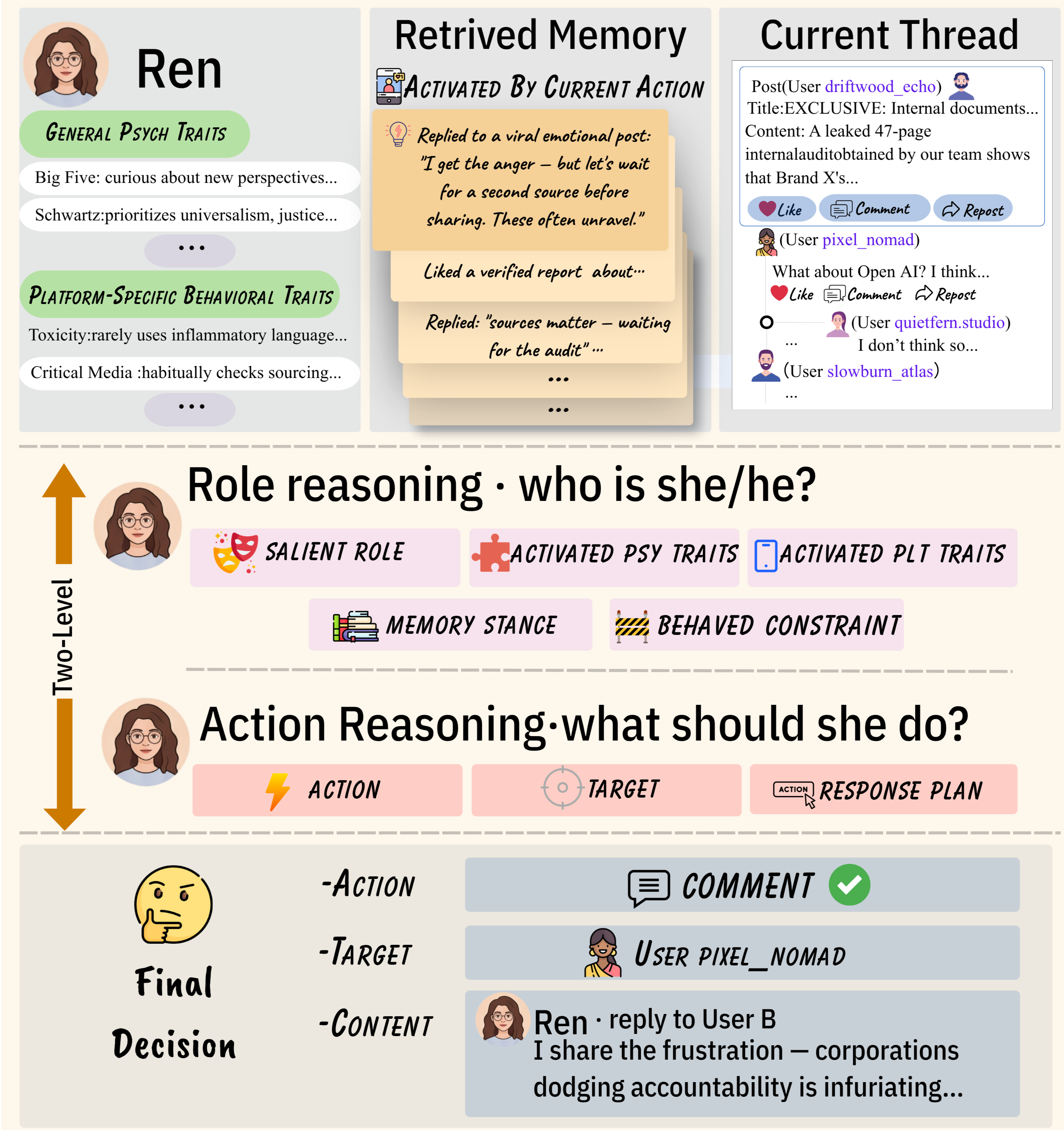}
	\vspace{-20pt}
	\caption{
		Motivation example of identity-driven social-media user simulation. 
		Given Ren's profile, retrieved memory, and current thread, the simulator first activates profile and memory evidence relevant to the thread, then selects the target before predicting the action and response.
	}
	\vspace{-6pt}
	\label{fig:motivation}
\end{figure}

Conventional character role-play usually starts from an explicit persona and evaluates whether generated dialogue remains persona-consistent \citep{xu-etal-2026-adamarp,10.1145/3770855.3818027, cheng-etal-2026-psymem}.
Social-media user simulation poses a different task: it infers observable platform behavior from a user's partial and noisy behavioral history together with the current thread context. The prediction covers both textual responses and structured decisions about target selection and the subsequent action type.
We formulate the task as predicting a user's observed engagement in a given post--comment context, including the target, subsequent action type, and optional textual response. 
We normalize observed interactions into \textsc{Comment}, \textsc{Like}, and \textsc{Repost}\footnote{Textual interactions such as replies and comments are mapped to \textsc{Comment}; resharing behaviors are mapped to \textsc{Repost}; and reaction-only behaviors are mapped to \textsc{Like}.}.
The logs record positive interactions but not impression-level exposure. Accordingly, the benchmark is conditional on a recorded engagement event, contains no explicit no-action label, and cannot distinguish unseen threads from threads that a user viewed but ignored. Within this scope, the task predicts the target, subsequent action, and optional text.

Existing LLM-based social simulation systems have explored information diffusion, online conversation, and social media event modeling \citep{park2023generative,gao2024large,hu2024laidsim,liu2025mosaic,jeon2025simulating,ji2026ripplemem}. 
These studies show that LLM agents can generate natural-language interactions and support controlled simulation of social processes. 
However, they often rely on coarse personas, demographic attributes, or manually specified agent descriptions, and are less focused on predicting a concrete user's platform behavior from long-horizon behavioral histories and situated thread context.

To address this, we introduce \textbf{PsyMemBehav}, a structured framework that predicts social media behavior by integrating user profiles, longitudinal memories, and thread context. The framework maps this evidence to behavior through two sequential steps: identity activation, which selects relevant traits and memories for the current thread, and action decision, which identifies the target and predicts the subsequent action and response. Furthermore, we align the complete reasoning-action trajectories via rubric-guided preference optimization. This ensures the simulator optimizes not just for response fluency, but also for action fidelity, memory grounding, and behavioral consistency.

To evaluate this, we construct a large-scale dataset PsyMemBehav-bench, containing 5.57M behaviors from 5,862 users across Reddit, Bluesky, and Twitter. Each sample of PsyMemBehav-bench integrates 41 psychology-informed descriptors, retrieved history, context, and observed behavior. This enables the joint evaluation of structured decision-making and text generation, moving beyond traditional text-only role-playing.

We evaluate PsyMemBehav at the individual level by matching held-out user interactions and at the thread level by comparing aggregate comment-section statistics in observed post contexts. In summary, our contributions are as follows:

\textbf{1)} We propose \textbf{PsyMemBehav}, a novel framework that models conditional social-media behavior through a two-stage structured decision process: \textit{identity activation} and \textit{action decision}. By dynamically integrating 41-dimensional psychology-informed user profiles and retrieved longitudinal memories with thread context, our approach moves beyond flat text generation to produce grounded, context-aware behavioral predictions.

\textbf{2)} We introduce a tailored alignment strategy combining Reverse-CoT supervision and rubric-guided DPO to optimize complete reasoning-action trajectories. This ensures the simulator is optimized not merely for response fluency, but critically for action fidelity (e.g., predicting \textsc{Comment}, \textsc{Like}, or \textsc{Repost} with conditional text generation), robust memory grounding, and strict persona consistency.

\textbf{3)} 
On the large-scale dataset PsyMemBehav-bench, we conduct comprehensive evaluations on both held-out user behavior and  observed-thread continuation. The resulting PsyMemBehav-27B ranks first in 11 of 12 individual metrics, achieving a 9.95\% relative improvement in average response quality over GPT-5.4, while yielding the lowest aggregate group-level deviation. Extensive ablation studies further validate the critical contributions of both training stages, profile dimensions, and decision components.

\section{Related Work}

\subsection{Social-Media Account Simulation}

LLM-based agents have been used to simulate human-like behavior in role-playing, interactive environments, and social settings \citep{park2023generative,argyle2023out,gao2024large}. Prior work shows that LLMs can maintain personas, imitate conversational styles, and produce role-consistent responses \citep{shao2023characterllm,wang2023rolellm,tu2024charactereval}. In parallel, user modeling studies show that digital traces reveal stable attributes such as personality, preferences, demographics, and behavioral tendencies \citep{kosinski2013private,youyou2015computer,hovy2015tagging}, motivating personalized behavior generation from historical traces.

Social-media account simulation, however, goes beyond generic persona imitation. Users do not only generate text, but also select interaction targets within a thread and choose platform-specific actions such as replying, quoting, reposting, liking, or ignoring. Although recent work explores LLM agents for interactive social behavior and social simulation \citep{ren2024bases,zhou2024misleading}, existing approaches often focus on generic personas, aggregate populations, or domain-specific simulated users rather than real accounts grounded in historical traces and platform-specific action spaces. Our work addresses this gap by modeling real accounts with psychological traits, platform-specific behavioral traits, retrieved memory, and thread context.

\subsection{Rubric-Guided Alignment for Social-Behavior Simulation}

Preference learning aligns LLMs with desired behaviors through human or AI feedback and direct preference optimization \citep{ouyang2022training,bai2022constitutional,lee2023rlaif,rafailov2023direct}, typically targeting general objectives such as helpfulness, harmlessness, or instruction following. Social-media simulation instead requires faithfulness to user profiles, grounding in memory, platform-appropriate actions, thread-context coherence, and social plausibility.

LLM-as-a-judge methods enable evaluation with explicit criteria rather than a single overall score \citep{zheng2023judging,liu2023geval,kim2023prometheus}. Building on this idea, we construct rubric-guided preference pairs over complete reasoning-action trajectories, aligning both intermediate reasoning and resulting platform behavior rather than only final textual responses.

\subsection{Group-Level Thread Simulation}

Agent-based modeling studies how individual behaviors produce collective phenomena such as diffusion, coordination, polarization, and network formation \citep{epstein1999agent,macy2002factors,lazer2009computational}. LLM agents extend this tradition with language-based reasoning and context-sensitive interaction policies \citep{park2023generative,gao2024large}, while introducing challenges such as unstable identities, unrealistic information access, and weak behavioral grounding.

Recent work argues that social simulation should be evaluated beyond single-turn plausibility, including interaction dynamics, stance distributions, network structures, conversational diversity, and thread-level realism \citep{zhou2024misleading,zhang2025gas3}. We therefore first improve account-level identity-to-action modeling through psychology-informed reasoning and rubric-guided DPO, and then evaluate whether these improvements transfer to group-level thread simulations.

\section{Methodology}
\label{sec:method}

\subsection{Problem Formulation and Overview}

We formulate social-media user simulation as a time-anchored multi-action prediction problem. 
For a user $u$ at time $t$, the simulator observes
\vspace{-0.1em}
\[
x_t = (P_u, M_{u,t}, T_t, C_t),
\vspace{-0.1em}
\]
where $P_u$ is the user profile, $M_{u,t}$ is retrieved behavioral memory before time $t$, $T_t$ is the current thread context, and $C_t$ is the set of candidate targets, i.e.\ the visible post or comments the user may interact with at $t$. 
The simulator predicts a complete platform behavior
\vspace{-0.1em}
\[
y_t = (c_t, a_t, r_t),
\vspace{-0.1em}
\]
where $c_t \in C_t$ is the selected target, $a_t \in \mathcal{A}$ the subsequent action, and $r_t$ the generated comment when required. 
We normalize the heterogeneous behaviors in raw logs into three visible actions, $\mathcal{A}=\{\textsc{Comment}, \textsc{Like}, \textsc{Repost}\}$: textual engagement (comments, replies) maps to \textsc{Comment}, reaction-only behavior to \textsc{Like}, and resharing (including quote reposts) to \textsc{Repost}; original posts and past interactions serve as evidence rather than prediction labels.

\begin{figure*}[t]
	\centering
	\includegraphics[width=\textwidth]{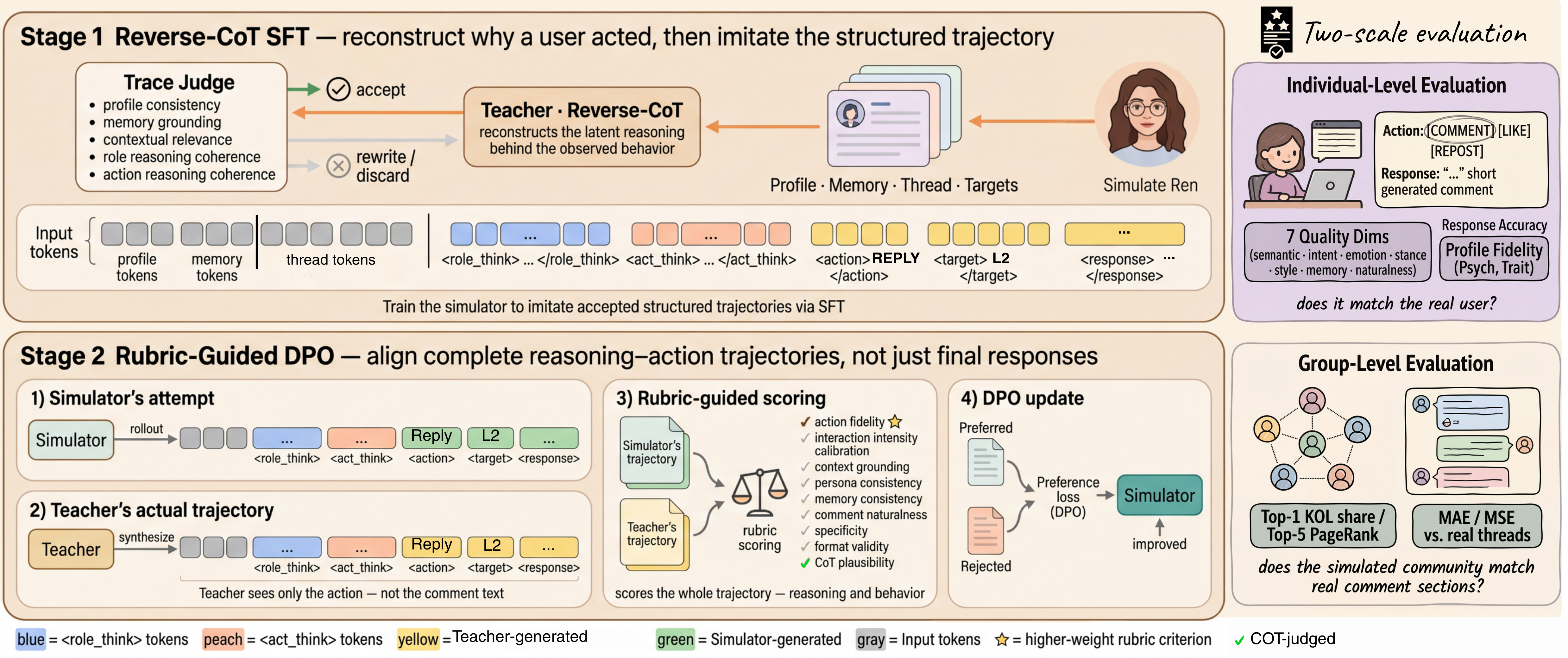}
	\vspace{-20pt}
	\caption{
		Overview of the proposed two-stage training framework. Stage 1 reconstructs reverse-CoT traces from observed behavior and trains the simulator by supervised fine-tuning on accepted trajectories. Stage 2 applies rubric-guided DPO to align complete reasoning-action trajectories with social-behavior criteria. The right panel summarizes the individual- and group-level evaluation.
	}
	\label{fig:framework}
\end{figure*}

This requires more than fluent response generation: conditional on a recorded engagement event, the simulator must first select a target, then choose Comment, Like, or Repost and generate text when required. 
Because behavior is context-dependent, a plausible comment can still be wrong if the user would target a different item or act differently. We address this with two design choices. 
First, \emph{two-level identity-to-action reasoning} situates the user in the current thread before deciding the platform behavior (\S\ref{sec:reasoning}). 
Second, \emph{two-stage trajectory training} optimizes complete reasoning-action trajectories rather than only final responses, through reverse-CoT supervision and rubric-guided preference optimization (\S\ref{sec:training}). Figure~\ref{fig:framework} gives an overview.

\subsection{Two-Level Identity-to-Action Reasoning}
\label{sec:reasoning}

Each simulation instance is modeled as two ordered decisions: the evidence relevant to the user in the current thread is identified first, and the corresponding platform behavior is predicted second.
This separation reflects the difference between a broad, history-derived profile and a context-specific action, while making the intermediate evidence selection explicit.
A direct mapping from $x_t$ to $y_t$ remains possible, but it does not expose separate supervision for evidence selection and action decision.
We therefore insert an intermediate reasoning trace
\vspace{-0.3em}
\[
x_t \rightarrow z_t \rightarrow y_t,
\qquad
z_t = (z_t^{\mathrm{id}}, z_t^{\mathrm{act}}),
\vspace{-0.3em}
\]
with identity activation $z_t^{\mathrm{id}}$ (which profile and memory evidence is relevant here) and action decision $z_t^{\mathrm{act}}$ (which target, action, and response follow from that evidence).

\paragraph{Psychology-informed representation.}
We organize user evidence into two complementary groups:
\vspace{-0.3em}
\[
P_u = (P_u^{\mathrm{gen}}, P_u^{\mathrm{plat}}),
\vspace{-0.3em}
\]
where $P_u^{\mathrm{gen}}$ contains 22 dimensions spanning Big Five personality, Schwartz values, moral foundations, and social dominance orientation \citep{goldberg1990alternative,schwartz1992universals,graham2013moral,pratto1994social}, while $P_u^{\mathrm{plat}}$ contains 19 dimensions spanning emotional tendencies, network and social capital, media literacy, and online behavior \citep{katz1955personal,ellison2007benefits,guess2020simple,wulczyn2017exmachina}. These 41 variables are LLM-inferred, history-derived descriptors for behavior prediction, not validated psychometric or clinical measurements. General traits encode cross-context dispositions, whereas platform traits encode online interaction tendencies, allowing identity activation to select the evidence relevant to a specific thread. Profiles are built once from each user's pre-split history, $P_u=g_{\mathrm{prof}}(\mathcal{H}_u^{\mathrm{train}})$, and reused for held-out instances to prevent post-split leakage. The complete taxonomy and annotation prompt are in the appendix.

\paragraph{Memory retrieval.}
We use the current visible thread as a lexical query, rank each strictly pre-time historical behavior with BM25 \citep{robertson2009bm25}, and retain the top five as $M_{u,t}$. The score uses neither dense similarity nor an additional recency prior. In the retrieval-size analysis, both 9B and 27B models peak at $k=5$; fewer memories omit relevant history, whereas more introduce redundancy or noise (see the appendix).

\paragraph{Identity activation and target-first action decision.}
Given the profile and retrieved memory, \emph{identity activation} selects and summarizes the profile traits, historical behaviors, stance evidence, and behavioral constraints relevant to the current thread.
Conditioned on this evidence summary, \emph{action decision} selects the target before predicting the action and response plan.
The decomposition provides explicit supervision for distinguishing \textsc{Comment}, \textsc{Like}, and \textsc{Repost} instead of treating every interaction as text generation.
Together they yield the complete trajectory
\vspace{-0.3em}
\[
\tau_t = (z_t, y_t) = (z_t^{\mathrm{id}}, z_t^{\mathrm{act}}, c_t, a_t, r_t).
\vspace{-0.3em}
\]
The full schema and a worked example are in the appendix.

\subsection{Two-Stage Training}
\label{sec:training}

We train over the trajectory $\tau_t$ in two stages (Figure~\ref{fig:framework}): supervised fine-tuning on reconstructed traces, then rubric-guided preference optimization over complete trajectories.

\paragraph{Stage 1: Reverse-CoT supervision.}
Behavior logs provide observed actions and comments but no explicit annotation of the underlying decision process. Direct output supervision therefore leaves the intermediate mapping from user evidence to platform behavior implicit. We reconstruct the missing reasoning after the fact: given $(x_t,y_t^*)$, a teacher produces a posterior rationale constrained by the observed behavior, $z_t^*=f_{\mathrm{teacher}}(x_t,y_t^*)$. 
Since $z_t^*$ is generated after the behavior, we treat it not as
ground-truth cognition but as structured supervision for learning the
mapping from user evidence to observed behavior. Following
behavior-conditioned user modeling paradigms
\citep{zhang2025llmpowered}, the reconstructed trace connects the user
profile, retrieved memory, and thread context with the observed action.
It is therefore intended to provide a plausible and behavior-consistent
reasoning structure, rather than to recover a unique latent cognitive
process. The observed behavior is used only during trace construction
and is unavailable at inference time.
A validation judge $J_{\mathrm{trace}}$ (checking profile consistency, memory grounding, contextual relevance, identity-activation coherence, and action-decision coherence; Figure~\ref{fig:framework}) keeps only consistent traces,
\vspace{-0.3em}
\[
\mathcal{D}_{\mathrm{SFT}}
=
\{(x_t,z_t^*,y_t^*) \mid J_{\mathrm{trace}}(x_t,z_t^*,y_t^*)=1\}_{t=1}^{N},
\vspace{-0.3em}
\]
and we minimize the negative log-likelihood of the full structured output,
\vspace{-0.3em}
\[
\mathcal{L}_{\mathrm{SFT}}
=
-\mathbb{E}_{(x,z^*,y^*)\sim\mathcal{D}_{\mathrm{SFT}}}
\log \pi_{\theta}(z^*,y^*\mid x),
\vspace{-0.3em}
\]
initializing the simulator to emit identity activation, action decision, and behavior in one trajectory. Judge criteria, the form of $J_{\mathrm{trace}}$, and prompts are in the appendix.

\paragraph{Stage 2: Rubric-guided trajectory DPO.}
SFT yields well-formed trajectories but does not say which of several fluent ones is the more faithful simulation. 
We therefore optimize preferences over complete trajectories. 
The rubric itself is induced rather than hand-specified. We sample trajectory pairs and ask a judge to write free-form critiques explaining why one trajectory is a less faithful simulation than the other; an LLM then consolidates the full set of critiques into ten criteria that together account for the observed weaknesses, which the human annotators verify for coverage and non-redundancy. 
Grounding the evaluation axes in the weaknesses simulators actually exhibit, rather than in a priori intuitions, makes the rubric an explanatory cover of the failures rather than an arbitrary checklist. 
Once induced, the rubric is frozen and applied uniformly during preference optimization.
For each input we compare a trajectory rolled out by the SFT model against a reference synthesized from the observed behavior, scoring both with a fixed ten-criterion social-behavior rubric (action fidelity, interaction-intensity calibration, context grounding, persona/memory/stance consistency, comment naturalness, specificity, format validity, and evidence-to-action coherence; full definitions in the appendix).
Two choices matter. 
First, since action selection is central, action fidelity is up-weighted,
\vspace{-0.6em}
\[
S(\tau) = \frac{2s_1(\tau)+\sum_{j=2}^{10}s_j(\tau)}{11},
\vspace{-0.6em}
\]
so the action is never outweighed by surface fluency. 
Second, the judge sees only the ground-truth action label $a_i^*$, \emph{not} the gold comment, forcing it to assess whether each trajectory chooses an appropriate action and uses evidence faithfully rather than matching reference wording. 
To avoid spurious supervision, we retain a pair only when the judge is confident, the score margin exceeds a threshold, the trajectories are not near-duplicates, and the preference is not explained by length.
Algorithm~\ref{alg:pref} summarizes the full construction.

Given the resulting $\mathcal{D}_{\mathrm{DPO}}=\{(x_i,\tau_i^+,\tau_i^-)\}_{i=1}^{M}$, we optimize a DPO objective~\citep{rafailov2023direct} over \emph{complete} trajectories: the preference margin scores the full $(z^{\mathrm{id}}_t, z^{\mathrm{act}}_t, c_t, a_t, r_t)$ under $\pi_\theta$ relative to the SFT reference $\pi_{\mathrm{ref}}$ with strength $\beta$, aligning both the intermediate reasoning and the platform behavior with the rubric. The complete rubric, trace-validation criteria, and judging prompts are in the appendix.

\begin{algorithm}[t]
	\caption{Trajectory Preference Construction}
	\label{alg:pref}
	\begin{algorithmic}[1]
		\REQUIRE input $x_i$; SFT policy $\pi_{\mathrm{SFT}}$; observed behavior $y_i^*$ with action label $a_i^*$; rubric score $S(\cdot)$
		\STATE $\tau_i^{\mathrm{gen}} \sim \pi_{\mathrm{SFT}}(\cdot \mid x_i)$ \hfill\COMMENT{simulator rollout}
		\STATE $\tau_i^{\mathrm{ref}} \gets (z_i^{\mathrm{ref}}, y_i^*)$ \hfill\COMMENT{synthesize reference from observed behavior}
		\STATE score $\tau_i^{\mathrm{gen}}, \tau_i^{\mathrm{ref}}$ with $S(\cdot)$, exposing $a_i^*$ to the judge but \emph{not} the gold comment
		\STATE $(\tau_i^+, \tau_i^-) \gets$ higher- / lower-scoring trajectory
		\IF{pair passes confidence, margin, duplication, and length filters}
		\STATE add $(x_i, \tau_i^+, \tau_i^-)$ to $\mathcal{D}_{\mathrm{DPO}}$
		\ENDIF
	\end{algorithmic}
\end{algorithm}

\begin{table*}[t]
	\centering
	\footnotesize 
	\setlength{\tabcolsep}{2.8pt}
	\renewcommand{\arraystretch}{1.0}
		\begin{tabular}{
				l
				cccccccccccc
			}
			\toprule
			\textbf{Models}
			& \multicolumn{10}{c}{\textbf{Behavior \& Response Quality}}
			& \multicolumn{2}{c}{\textbf{Profile Fidelity}} \\
			\cmidrule(lr){2-11} \cmidrule(lr){12-13}
			& \textbf{Act.}
			& \textbf{Tgt.}
			& \textbf{Sem.}
			& \textbf{Intent}
			& \textbf{Emo.}
			& \textbf{Stance}
			& \textbf{Style}
			& \textbf{Memory}
			& \textbf{Natural.}
			& \textbf{Avg.}
			& \textbf{Psych}
			& \textbf{Trait} \\
			\midrule
			GPT-oss-120B
			& 54.63 & 37.35 & 34.49 & 41.50 & 43.39 & 51.39 & 37.70 & 52.14 & 45.95 & 42.92
			& 73.12 & 73.78 \\
			GPT-5
			& 61.80 & 58.80 & 46.99 & 54.44 & 54.22 & 65.12 & 46.36 & 60.79 & 58.57 & 54.12
			& 76.52 & 75.23 \\
			Gemini-2.5-Pro
			& 63.43 & 49.28 & 41.43 & 44.45 & 45.97 & 58.07 & 42.83 & 54.91 & 61.28 & 49.84
			& 72.86 & 71.85 \\
			GPT-5.4
			& 66.12 & 59.46 & 45.05 & 54.21 & 56.36 & \textbf{69.25} & 45.49 & 58.19 & 62.94 & 54.28
			& 76.59 & 75.11 \\
			Claude-Sonnet-4.6
			& 63.67 & 57.23 & 41.57 & 49.83 & 51.84 & 62.62 & 43.61 & 55.58 & 60.40 & 50.75
			& 77.09 & 71.39 \\
			Qwen3.7-Max Preview
			& 62.02 & 54.01 & 40.85 & 49.47 & 52.16 & 63.76 & 45.18 & 58.98 & 56.60 & 51.16
			& 74.73 & 72.87 \\
			Gemini-3.1-Pro-Preview
			& 62.81 & 51.61 & 37.53 & 46.26 & 50.24 & 60.77 & 48.12 & 57.66 & 60.02 & 50.26
			& 75.91 & 73.72 \\
			DeepSeek-V4-Pro
			& 64.32 & 47.01 & 36.62 & 45.15 & 47.36 & 59.94 & 40.70 & 52.96 & 55.19 & 46.86
			& 75.63 & 71.35 \\
			\midrule
			Qwen3.5-9B
			& 51.61 & 38.79 & 26.89 & 33.42 & 38.95 & 48.28 & 33.09 & 39.55 & 36.12 & 35.60
			& 70.77 & 70.25 \\
			\rowcolor{oursblue}
			PsyMemBehav 9B
			& 59.52 & 48.22 & 36.87 & 46.58 & 48.71 & 58.54 & 40.25 & 51.64 & 58.90 & 48.78
			& 75.33 & 74.43 \\
			\midrule
			Qwen3.5-27B
			& 56.57 & 47.90 & 33.87 & 39.84 & 42.23 & 49.43 & 40.30 & 54.52 & 49.62 & 44.26
			& 72.23 & 71.02 \\
			\rowcolor{oursblue}
			\textbf{PsyMemBehav 27B}
			& \textbf{71.40} & \textbf{65.34} & \textbf{50.35} & \textbf{55.49} & \textbf{58.89} & 68.89 & \textbf{53.57} & \textbf{64.19} & \textbf{66.39} & \textbf{59.68}
			& \textbf{80.38} & \textbf{78.25} \\
			\bottomrule
		\end{tabular}
	\vspace{-6pt}
	\caption{
		Individual-level user simulation results. Blue rows denote our proposed PsyMemBehav models. 
		\textbf{Act.} denotes action accuracy over \textsc{Comment}, \textsc{Like}, and \textsc{Repost}; 
		\textbf{Tgt.} denotes target accuracy. The seven response-quality dimensions are judged by Gemini-3-Flash. 
		\textbf{Psych} and \textbf{Trait} are profile-fidelity scores computed by re-annotating generated behaviors and comparing them with the original user profile using $100 \times (1-\mathrm{NED})$. 
		Higher is better for all metrics. Best results are bolded.
	}
	\vspace{-5pt}
	\label{tab:individual_results}
\end{table*}

\begin{table}[t]
\centering
\footnotesize
\setlength{\tabcolsep}{4.2pt}
\renewcommand{\arraystretch}{0.96}
\begin{tabular}{lrrr}
\toprule
\textbf{Dimension} & \textbf{Human mean} & \textbf{Gemini} & \textbf{Human$-$Gemini} \\
\midrule
Semantic        & 44.22 & 41.98 & $+2.24$ \\
Intent          & 48.46 & 50.84 & $-2.38$ \\
Emotion         & 50.94 & 54.54 & $-3.60$ \\
Stance          & 52.31 & 56.51 & $-4.20$ \\
Style           & 55.79 & 54.38 & $+1.41$ \\
Profile/memory  & 61.29 & 65.02 & $-3.73$ \\
Naturalness     & 67.77 & 64.92 & $+2.85$ \\
\midrule
\textbf{Overall} & \textbf{54.40} & \textbf{55.46} & $\mathbf{-1.06}$ \\
\bottomrule
\end{tabular}
\vspace{-8pt}
\caption{Human and Gemini mean scores on 150 instances from 30 anonymized user IDs (five per ID); differences are human minus Gemini.}
\label{tab:human_gemini_agreement}
\end{table}

\section{Experimental Setup}
\label{sec:experimental_setup}

To evaluate the effectiveness of the proposed PsyMemBehav framework, we conduct experiments on the social-media user simulation task defined in Section~\ref{sec:method}. The model first selects the target, then predicts \textsc{Comment}, \textsc{Like}, or \textsc{Repost} and any required comment text.

\subsection{Base Models}

We conduct experiments on two backbone models: Qwen3.5-9B and Qwen3.5-27B. The 9B model is used to examine whether the proposed framework remains effective under a smaller simulator, while the 27B model serves as the stronger backbone for our main experiments.
Both models are trained under the same two-stage pipeline, resulting in two models: PsyMemBehav 9B and PsyMemBehav 27B.

\subsection{Baselines}
We compare PsyMemBehav 9B and PsyMemBehav 27B against the following baselines: GPT-5.4 \citep{openai2026gpt54}, Gemini-2.5-Pro \citep{comanici2025gemini}, Claude-Sonnet-4.6 \citep{anthropic2025claude}, Qwen3.7-Max \citep{yang2025qwen3}, and DeepSeek-V4-Pro \citep{xu2026deepseek}. 
Additionally, Qwen3.5-9B and Qwen3.5-27B \citep{yang2025qwen3} are included as open-weight baselines.
All models receive the same full user profile, BM25-retrieved pre-event memories, thread context, and candidate targets. They use the unified end-to-end prompt in the appendix,  which selects a target before predicting the action and branch-specific text; gold target and action labels are withheld.

\subsection{Evaluation Metrics}

We evaluate the models at both individual and group levels.

\paragraph{Individual-level evaluation.}
We report two behavior-level metrics. 
\textbf{Target Accuracy} measures whether the model selects the correct interaction target among the candidate posts and comments. 
\textbf{Action Accuracy} measures whether the model chooses the correct platform action---\textsc{Comment}, \textsc{Like}, or \textsc{Repost}---for the selected target.

For generated comments, we use Gemini-3-Flash as an automated judge to score seven quality dimensions: {semantic alignment}, {intent alignment}, {emotional consistency}, {stance consistency}, {style fidelity}, {persona memory}, and {naturalness}. We report both individual dimension scores and their average. The complete scoring prompt appears in the appendix.

\paragraph{Group-level simulation.}
To evaluate multi-user thread continuation, we place profiled simulators in the same observed thread seed and roll out their target-first actions. Following prior work that evaluates collective simulations through interaction, stance, and influence structure \citep{zhou2024misleading,liu2025mosaic}, we report four complementary metrics. \emph{Reply Reciprocity} is the share of interacting dyads with mutual directed replies, a standard property of directed networks \citep{garlaschelli2004reciprocity}. \emph{Stance deviation} is the $\ell_1$ distance between simulated and real \textsc{Pro}, \textsc{Anti}, and \textsc{Neutral} distributions, using the conventional favor/against/neither stance formulation \citep{mohammad2016semeval}. \emph{Top1 KOL} is the reply share received by the most-addressed user, operationalizing key-opinion-leader concentration \citep{katz1955personal}; \emph{Top5 PageRank} is the normalized PageRank mass of the five highest-ranked users \citep{brin1998anatomy}. MAE and MSE summarize absolute and squared deviations across thread-level statistics. Detailed formulas are in the appendix.

\subsection{Dataset and Data Split}

We construct PsyMemBehav-bench from Reddit, Bluesky, and Twitter through official APIs. It contains 5.57M behaviors from 5,862 real users (approximately 950 per user), with usernames replaced by randomized hash identifiers. Platform events are normalized as \textsc{Comment}, \textsc{Like}, or \textsc{Repost}. Each instance combines a two-group profile, five BM25-retrieved pre-event memories, thread context, and candidate targets with the observed target, action, and response. A user-disjoint chronological split assigns 5,000 users to training and 862 to validation and testing; held-out users are unseen, and all profile and memory evidence predates prediction. Collection favors accounts with rich observable histories. Thus, the scale denotes longitudinal density rather than platform-wide representativeness and may overrepresent active users while underrepresenting low-activity, private, or new users.
Among closely related studies reporting real-user counts, FineRob uses 1,866 users and 78.6K records (approximately 126 histories per user), whereas MOSAIC uses 204 real users and 30 posts and GA-S$^3$ models group agents over 30 events. PsyMemBehav therefore provides larger and denser longitudinal evidence for user-level target--action--response modeling, not a claim of platform-wide coverage (see the appendix).

\begin{table}[t]
	\centering
	\small
	\setlength{\tabcolsep}{3.5pt}
	\begin{tabular}{lccccc}
		\toprule
		\textbf{Variant} & \textbf{Act.} & \textbf{Tgt.} & \textbf{Avg.} & \textbf{Psych} & \textbf{Trait} \\
		\midrule
		\multicolumn{6}{l}{\textit{Training ablation}} \\
		Base      & 51.61 & 38.79 & 35.60 & 70.77 & 70.25 \\
		CoT-P     & 47.31 & 35.90 & 33.92 & 69.69 & 68.81 \\
		Out-SFT   & 49.23 & 40.30 & 39.70 & 71.73 & 70.34 \\
		R-CoT     & 53.61 & 46.64 & 43.37 & 74.58 & \textbf{75.79} \\
		R-CoT+R-DPO & \textbf{59.52} & \textbf{48.22} & \textbf{48.78} & \textbf{75.33} & 74.43 \\
		\midrule
		\multicolumn{6}{l}{\textit{Component ablation under R-CoT}} \\
		Full      & \textbf{53.61} & \textbf{46.64} & \textbf{43.37} & \textbf{74.58} & \textbf{75.79} \\
		w/o Gen.  & 52.48 & 43.77 & 39.66 & 70.71 & 74.69 \\
		w/o Plat. & 52.39 & 42.32 & 41.13 & 73.11 & 71.45 \\
		w/o IdAct & 51.27 & 41.03 & 38.77 & 69.43 & 72.50 \\
		w/o ActDec & 50.47 & 40.62 & 39.37 & 69.89 & 71.83 \\
		\bottomrule
	\end{tabular}
    \vspace{-6pt}
	\caption{
		Ablation results on Qwen3.5-9B. 
		Act. is action accuracy; Tgt. is target accuracy; Avg. is the average score over seven response-quality dimensions. 
		Psych and Trait are profile-fidelity scores for general psychological traits and platform-specific traits.
	}
	\label{tab:ablation}
\end{table}

\begin{table*}[t]
	\centering
	\small
	\setlength{\tabcolsep}{6pt}
	\renewcommand{\arraystretch}{1.0}
		\begin{tabular}{lcccccc}
			\toprule
			\textbf{Model}
			& \textbf{Reciprocity}
			& \textbf{Stance Dev. $\downarrow$}
			& \textbf{Top1 KOL}
			& \textbf{Top5 PR}
			& \textbf{MAE $\downarrow$}
			& \textbf{MSE $\downarrow$} \\
			\midrule
			Real
			& 0.043 & 0.000 & 0.213 & 0.596 & 0.000 & 0.000 \\
			\midrule
			\textbf{PsyMemBehav 27B}
			& \textbf{0.121} & \textbf{0.155} & 0.232 & {0.524} & \textbf{0.081} & \textbf{0.009} \\
			
			GPT-5.4
			& 0.182 & 0.175 & 0.197 & \textbf{0.529} & 0.099 & 0.014 \\
			
			Claude-Sonnet-4.6
			& 0.313 & 0.322 & 0.191 & 0.472 & 0.185 & 0.048 \\
			
			GPT-5
			& 0.375 & 0.496 & \textbf{0.226} & 0.507 & 0.233 & 0.091 \\
			
			Qwen3.7-Max-Preview
			& 0.458 & 0.452 & 0.131 & 0.483 & 0.266 & 0.099 \\
			
			DeepSeek-V4-Pro 
			& 0.444 & 0.649 & 0.136 & 0.448 & 0.319 & 0.152 \\
			\bottomrule
		\end{tabular}
	\vspace{-6pt}
	\caption{
		Thread-level continuation results. We use \textbf{PsyMemBehav 27B} as our simulator and compare its generated interaction threads with those produced by strong general models. \textbf{Real} denotes statistics computed from real comment sections. \textbf{Reciprocity} measures bidirectional reply interactions. \textbf{Stance Dev.} measures deviation from the real stance distribution. \textbf{Top1 KOL} and \textbf{Top5 PR} measure the concentration of influence in the simulated interaction graph. \textbf{MAE} and \textbf{MSE} summarize the overall deviation from real group-level statistics. Lower is better for metrics marked with $\downarrow$; for other metrics, closer to the real distribution indicates better simulation fidelity.
	}
	\vspace{-5pt}
	\label{tab:group_simulation}
\end{table*}

\section{Experimental Results and Analyses}
\label{sec:results} 

\subsection{Individual-Level Results}
Table~\ref{tab:individual_results} reports target and action accuracy, seven response-quality dimensions, and profile fidelity, computed by re-annotating generated behavior and applying $100 \times (1-\mathrm{NED})$ to the original profile (see the appendix).

\textbf{PsyMemBehav 27B} ranks first in \textbf{11 of 12 metrics}, with target accuracy of 65.34, action accuracy of 71.40, average response quality of 59.68, and profile-fidelity scores of 80.38 (Psych) and 78.25 (Trait). Relative to GPT-5.4, its average response quality rises from 54.28 to 59.68 (9.95\% relative), alongside higher target and action accuracy.

Training improves both backbones consistently: PsyMemBehav 9B raises average response quality from 35.60 to 48.78 over Qwen3.5-9B, and PsyMemBehav 27B raises it from 44.26 to 59.68 over Qwen3.5-27B, with concurrent gains in target accuracy, action accuracy, and profile fidelity.

At 27B, the complete training pipeline improves action accuracy by 14.83 points, target accuracy by 17.44, average response quality by 15.42, Psych fidelity by 8.15, and Trait fidelity by 7.23 over the matched Qwen3.5 backbone. At 9B, the corresponding gains are 7.91, 9.43, 13.18, 4.56, and 4.18 points. The only metric not led by PsyMemBehav 27B is stance consistency: its 68.89 is 0.36 points below GPT-5.4's 69.25. Gains therefore span decisions, responses, and profile preservation.

\paragraph{Evaluation robustness.}
We score the same anonymized outputs with Qwen3.5-27B, GPT-5.4, Gemini-2.5-Pro, and Claude-Sonnet-4.6 as independent judges. PsyMemBehav 27B ranks first under all four, exceeding the runner-up by 7.73, 3.03, 4.90, and 2.83 points, respectively. Agreement across judge families reduces the likelihood that the main ranking is driven by one evaluator's self-preference, although it does not by itself establish human validity. Full cross-judge scores are reported in the appendix.

Table~\ref{tab:human_gemini_agreement} calibrates the automated evaluator against human judgments on 150 instances from 30 anonymized user IDs (five per ID). The overall means differ by 1.06 points (54.40 vs. 55.46), and the signed dimension gaps are mixed rather than uniformly favoring one evaluator. This close aggregate alignment supports model-level comparison, while the stance (4.20) and emotion (3.60) gaps motivate dimension-specific caution; full analysis appears in Appendix.

\subsection{Ablation Study}
\label{sec:ablation}

Table~\ref{tab:ablation} separates training ablations from component ablations on Qwen3.5-9B. \textit{R-CoT/CoT-SFT} is supervised fine-tuning on accepted reverse-CoT traces without DPO, whereas \textit{R-DPO} is rubric-guided trajectory optimization.

Naive CoT prompting degrades performance, and output-only SFT improves response quality without improving action prediction. R-CoT instead improves target accuracy, action accuracy, and profile fidelity; adding R-DPO further improves most behavior and response metrics. Relative to Base, the complete \textbf{R-CoT+R-DPO} pipeline raises target accuracy from 38.79 to \textbf{48.22}, action accuracy from 51.61 to \textbf{59.52}, and average response quality from 35.60 to \textbf{48.78}.

The two stages contribute differently. Relative to Base, R-CoT adds 7.85 points in target accuracy, 2.00 in action accuracy, 7.77 in average response quality, 3.81 in Psych fidelity, and 5.54 in Trait fidelity. Adding R-DPO contributes a further 1.58, 5.91, 5.41, and 0.75 points to target accuracy, action accuracy, response quality, and Psych fidelity, while Trait fidelity decreases by 1.36. Thus, reverse-CoT accounts for most target and trait gains; DPO mainly strengthens action selection and response quality despite a small trait-preservation trade-off.

Removing either profile group degrades performance, confirming that general psychological and platform-specific evidence are complementary. Removing identity activation mainly reduces response quality, whereas removing action decision produces the lowest target and action accuracy, supporting their respective roles in evidence selection and target-first behavior prediction.

The component losses also reveal specialization. Removing identity activation causes the largest drop in average response quality ($-4.60$) and Psych fidelity ($-5.15$), whereas removing action decision causes the largest losses in target accuracy ($-6.02$) and action accuracy ($-3.14$). Platform-specific traits contribute more to target accuracy ($-4.32$) and Trait fidelity ($-4.34$) than general traits ($-2.87$ and $-1.10$), while general traits contribute more to response quality ($-3.71$ versus $-2.24$). This pattern matches the intended division of labor among evidence and reasoning components.

\subsection{Group-Level Thread Simulation}

Table~\ref{tab:group_simulation} compares PsyMemBehav 27B with strong general models on continuations of the same observed thread contexts. \textbf{PsyMemBehav 27B} obtains the \textbf{lowest stance deviation (0.155), MAE (0.081), and MSE (0.009)}, while its reciprocity (0.121) is closest to the real value (0.043). Its lower reciprocity error also avoids the over-interactive group-chat pattern produced by general-purpose agents. Relative to GPT-5.4, PsyMemBehav reduces stance deviation by 11.4\%, MAE by 18.2\%, and MSE by 35.7\%. GPT-5 and GPT-5.4 remain closer on Top1 KOL and Top5 PageRank, respectively, so PsyMemBehav does not lead every structural statistic. In direct pairwise evaluation, \textbf{PsyMemBehav} is preferred over GPT-5.4 in \textbf{64.0\%} of thread continuations versus 34.0\%, with 2.0\% ties (Appendix). These results concern continuation within observed threads, not platform-wide social dynamics.

PsyMemBehav's absolute reciprocity gap from the real threads is 0.078, compared with 0.139 for GPT-5.4, a 43.9\% reduction. For influence concentration, however, its Top1 KOL gap is 0.019 versus 0.013 for GPT-5, and its Top5 PageRank gap is 0.072 versus 0.067 for GPT-5.4. The strongest improvements therefore concern interaction reciprocity, stance composition, and aggregate fidelity, while centralization remains model-dependent.


\section{Conclusion and Limitations}
\label{sec:conclusion}
We present PsyMemBehav, combining 41-dimensional user descriptors, BM25 memories, target-first reasoning, reverse-CoT supervision, and rubric-guided DPO. PsyMemBehav 27B achieves the strongest individual results and lowest aggregate deviation in observed-thread continuation, supporting evidence-grounded user models for micro-level public-opinion simulation. Limitations include missing exposure/no-action labels, LLM-inferred rather than self-reported profiles, answer-conditioned reasoning traces, and aggregate rather than per-instance human validation. Future work should address these issues across broader settings.

\section*{Acknowledgments}
This work is supported by the National Natural Science Foundation of China (No. 62206259).

\bibliography{aaai2027}

\appendix
\section{User Representation and Memory Retrieval}
\label{app:user_representation}

This technical supplement provides additional details on the user representation, memory retrieval, reasoning traces, evaluation rubric, qualitative examples, and prompt templates used in PsyMemBehav.
We organize the supplement around the main components of the system: user profile construction, memory retrieval, trace construction, DPO preference construction, additional analyses, case studies, and prompts.

\subsection{Trait-based User Representation}
\label{app:trait_representation}
\label{app:traits}

We represent each user with two complementary profile groups:
\[
P_u=(P_u^{\mathrm{gen}},P_u^{\mathrm{plat}}).
\]
The general psychological profile $P_u^{\mathrm{gen}}$ captures relatively stable dispositions, including personality traits, values, moral preferences, and broad social orientation.
These dimensions describe how a user tends to interpret information, express disagreement, evaluate social issues, and maintain persistent preferences across topics.

The platform-specific profile $P_u^{\mathrm{plat}}$ captures online behavioral tendencies.
These dimensions describe how the same user tends to behave in social-media environments, including opinion leadership, social capital, critical media literacy, and toxicity-related interaction patterns.
This separation is important because two users may share similar psychological dispositions while expressing them differently on a platform: one may engage through careful correction, another through short ironic replies, and another through passive endorsement.

The general psychological profile $P_u^{\mathrm{gen}}$ comprises 22 dimensions across four categories.
\textit{Big Five personality} \citep{goldberg1990alternative} captures openness, conscientiousness, extraversion, agreeableness, and neuroticism.
\textit{Schwartz values} \citep{schwartz1992universals} cover ten motivational value types: self-direction, stimulation, hedonism, achievement, power, security, conformity, tradition, benevolence, and universalism.
\textit{Moral foundations} \citep{graham2013moral} include care/harm, fairness/cheating, loyalty/betrayal, authority/subversion, and sanctity/degradation.
\textit{Social dominance orientation} \citep{pratto1994social} provides two dimensions: SDO dominance and SDO anti-egalitarianism.

The platform-specific profile $P_u^{\mathrm{plat}}$ comprises 19 dimensions across four categories.
\textit{Emotional profile} covers expressiveness, stability, negativity bias, empathy, anxiety tendency, and anger tendency.
\textit{Network and social capital} \citep{katz1955personal,ellison2007benefits} captures opinion leadership, network centrality, and social capital.
\textit{Media literacy} \citep{guess2020simple} includes critical media literacy, source evaluation, and misinformation susceptibility.
\textit{Online behavior} \citep{wulczyn2017exmachina} covers posting frequency, reply tendency, repost tendency, like tendency, toxicity, trolling tendency, and bot-likeness.

The profile is produced from each user's pre-split history $\mathcal{H}_u^{\mathrm{train}}$ by an annotation model $g_{\mathrm{prof}}$.
To avoid leakage, only behaviors before the prediction split are used for profile construction.
Each stored profile contains numerical trait scores and evidence-grounded summaries.
The numerical scores provide compact control variables, while the summaries preserve interpretable evidence for generation and analysis.

\subsection{Memory Retrieval}
\label{app:memory_retrieval}

For each prediction instance at time $t$, PsyMemBehav retrieves a compact memory set $M_{u,t}$ from the user's previous behaviors:
\[
\mathcal{H}_u^{<t}=\{h_{u,1},h_{u,2},\ldots,h_{u,n}\}.
\]
We use BM25 lexical retrieval \citep{robertson2009bm25}. The current post together with its visible comment context forms the query $q_t$, and each historical behavior $h_{u,j}$ is represented as a candidate document $d_{u,j}$ using its available textual content. The BM25 relevance score is
\[
\begin{aligned}
s_{\mathrm{BM25}}(d_{u,j},q_t)
&= \sum_{w\in q_t} \operatorname{IDF}(w)
\frac{f(w,d_{u,j})(k_1+1)}{D(w,d_{u,j})}, \\D(w,d_{u,j})
&= f(w,d_{u,j})
+ k_1\left(1-b+b\frac{|d_{u,j}|}{\mathrm{avgdl}}\right).
\end{aligned}
\]
where $f(w,d_{u,j})$ is the frequency of query term $w$ in document $d_{u,j}$, $|d_{u,j}|$ is the document length, $\mathrm{avgdl}$ is the average candidate-document length, and $k_1$ and $b$ are the standard BM25 term-saturation and length-normalization parameters. The final memory set keeps the top-ranked pre-time behaviors:
\[
M_{u,t}
=
\operatorname{TopK}_{h_{u,j}\in\mathcal{H}_u^{<t}}
s_{\mathrm{BM25}}(d_{u,j},q_t).
\]
We do not use dense embedding similarity or an additional recency-prior term in the retrieval score. Temporal validity is enforced by restricting the candidate pool to behaviors observed strictly before $t$. Unless otherwise specified, we use $k=5$ as the default retrieval size, balancing relevant historical evidence against redundant or noisy memories.

\subsection{User Profile Distribution}
\label{app:pca_density}

\begin{figure}[t]
\centering
\includegraphics[width=\columnwidth]{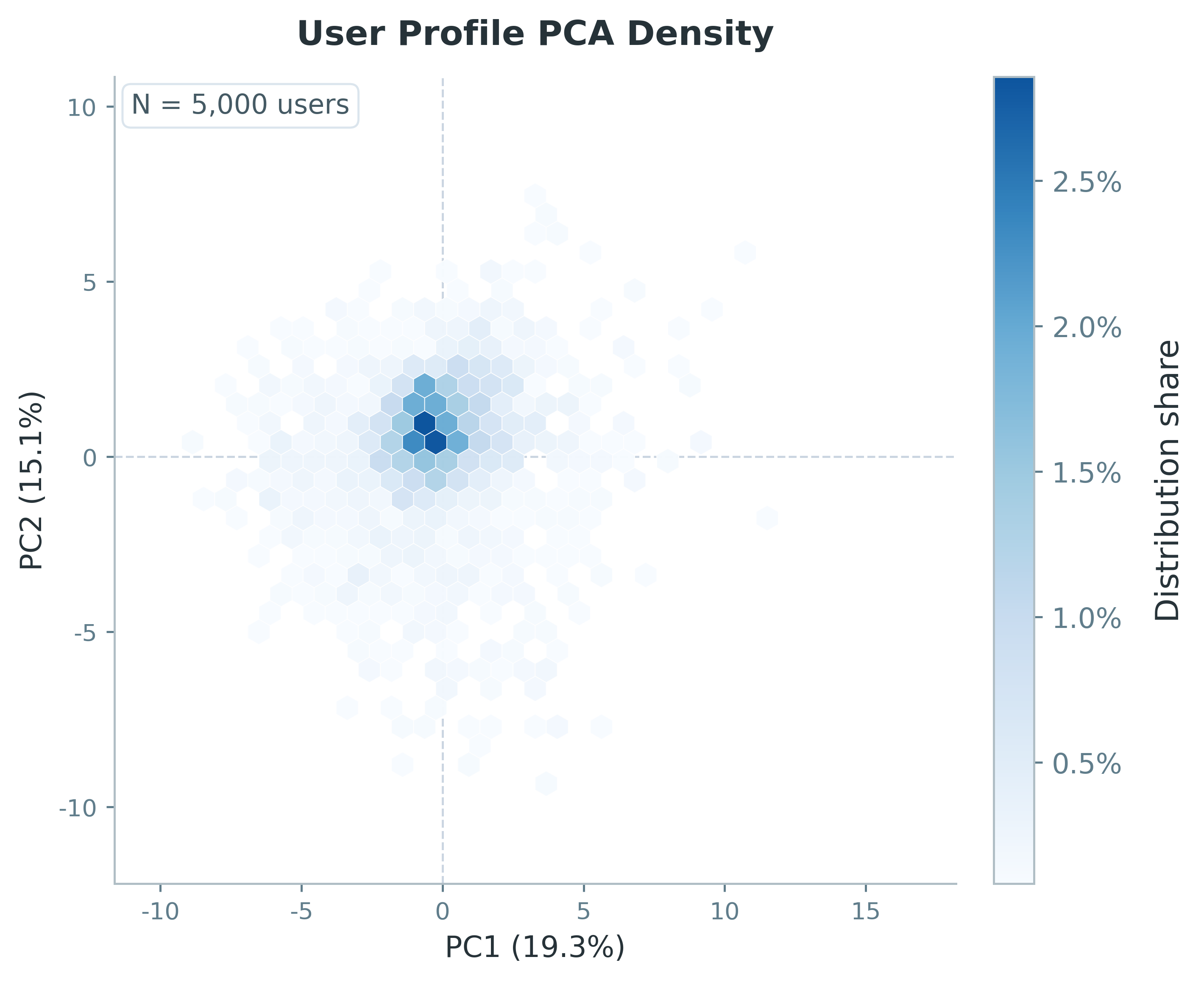}
\caption{
PCA density visualization of user profile representations.
Each hexagonal bin represents the local density of users in the projected two-dimensional profile space.
}
\label{fig:pca_density}
\end{figure}

To examine whether the constructed user profiles collapse into a narrow region or preserve meaningful diversity, we project the high-dimensional profile vectors into a two-dimensional space using principal component analysis.
Figure~\ref{fig:pca_density} shows that the profile distribution is compact but non-collapsed.
Most users are concentrated around a dense center, indicating shared behavioral regularities across the population.
At the same time, peripheral low-density regions indicate long-tail users with more distinctive profile combinations.
This supports the use of profile-conditioned simulation, since the population contains both central tendencies and individual heterogeneity.

\section{Reasoning Schema and Trace Construction}
\label{app:reasoning_schema}
\label{app:schema}
\label{app:revcot}

PsyMemBehav uses a two-level reasoning schema to connect user identity with concrete platform behavior.
The intermediate trace is represented as
\[
z_t=(z_t^{\mathrm{id}},z_t^{\mathrm{act}}),
\]
where $z_t^{\mathrm{id}}$ summarizes the situated user identity and $z_t^{\mathrm{act}}$ explains the selected target, the action conditioned on that target, and the resulting response when required.

The identity-activation component identifies the salient user role in the current context.
It includes activated psychological traits, activated platform-specific traits, retrieved memory evidence, likely stance, and behavioral constraints.
For example, a user may act as a skeptical fact-checker in one thread, a supportive community member in another, and an ironic commentator in a third.
The identity-activation stage is designed to capture this context-dependent but profile-consistent identity.

The action-decision component maps the situated evidence summary into a target-conditioned platform behavior:
\[
c_t\in C_t,\qquad
a_t\in\{\textsc{Comment},\textsc{Like},\textsc{Repost}\}.
\]
It first selects the interaction target $c_t$ and then predicts how to interact with that target through $a_t$. The visible response $r_t$ is planned only when the selected action is \textsc{Comment}.
For \textsc{Like} and \textsc{Repost}, the model returns the selected target and action without generating unnecessary visible text.

We construct reverse CoT traces by asking a teacher model to explain why the observed ground-truth action is plausible given the user profile, retrieved memory, and current thread.
The teacher is not asked to predict a new action; instead, it produces an inverse reasoning trace aligned with the known target and action.
This makes the reasoning trace a supervision signal for behavior explanation rather than a free-form alternative response.

A trace validator $J_{\mathrm{trace}}$ accepts a reconstructed trace only when it satisfies profile consistency, memory grounding, contextual relevance, role-reasoning coherence, and action-reasoning coherence:
\[
J_{\mathrm{trace}}(x_t,z_t,y_t^*)
=
\prod_{k=1}^{5}\mathbf{1}\big[c_k(x_t,z_t,y_t^*)\big]\in\{0,1\}.
\]
Only accepted traces are used as training supervision.

\section{Evaluation Rubric and DPO Preference Construction}
\label{app:rubric}

We use a ten-criterion rubric to evaluate simulated user behaviors and construct DPO preference pairs.
The rubric covers four aspects: action-level fidelity, user-profile alignment, content quality, and format/reasoning validity.
This design reflects the fact that social-media simulation is not only a text generation problem: the model must first select a valid interaction target, then choose how to interact with that target, and only then produce a response if the selected action requires visible text.

Action-level fidelity includes action accuracy, interaction-intensity calibration, and target/thread grounding.
User-profile alignment includes persona consistency, memory activation discipline, and stance consistency.
Content quality includes response naturalness and specificity.
Format/reasoning validity includes output format validity and CoT plausibility.

\begin{table*}[t]
\centering
\scriptsize
\setlength{\tabcolsep}{4pt}
\renewcommand{\arraystretch}{1.15}
\resizebox{\textwidth}{!}{
\begin{tabular}{
>{\raggedright\arraybackslash}p{2.8cm}
>{\raggedright\arraybackslash}p{3.3cm}
>{\raggedright\arraybackslash}p{11.5cm}
}
\toprule
\textbf{Category} & \textbf{Criterion} & \textbf{Definition} \\
\midrule

\multirow{3}{=}{\textbf{Action-level Fidelity}}
& \textbf{R1. Action Fidelity}
& Whether the candidate selects the correct high-level behavior type, such as \textsc{Comment}, \textsc{Like}, or \textsc{Repost}. \textsc{LikeThread} and \textsc{LikeBasic} are both treated as \textsc{Like}. \\

& \textbf{R2. Interaction Intensity}
& Whether the amount of engagement is appropriate. Overly strong interaction, such as writing a long comment when the user would only like, and under-interaction, such as producing a weak or incomplete response when a reply is expected, are penalized. \\

& \textbf{R3. Target Grounding}
& Whether the candidate understands the current post, comment tree, reply target, and local social context. \\

\midrule

\multirow{3}{=}{\textbf{User-profile Alignment}}
& \textbf{R4. Persona Consistency}
& Whether the behavior matches the user's profile, long-term interests, values, directness, social habits, and emotional style. \\

& \textbf{R5. Memory Discipline}
& Whether historical comments are used only when helpful. The model is penalized both for ignoring clearly relevant memories and for forcing irrelevant memories into the reasoning. \\

& \textbf{R6. Stance Consistency}
& Whether the stance is compatible with the user's prior positions and the current thread, without unsupported ideological invention or stance reversal. \\

\midrule

\multirow{2}{=}{\textbf{Content Quality}}
& \textbf{R7. Response Naturalness}
& Whether the visible response sounds like a real platform user rather than an assistant, summary, template, or overly polished explanation. \\

& \textbf{R8. Specificity}
& Whether the response addresses concrete contextual details rather than giving a generic or vague reply. \\

\midrule

\multirow{2}{=}{\textbf{Format and Reasoning}}
& \textbf{R9. Format Validity}
& Whether target, action, CoT, and visible response fields are complete, legal, parseable, and consistent with the candidate target list. \\

& \textbf{R10. CoT Plausibility}
& Whether the reasoning plausibly explains the action and response without over-rationalization, fabricated private psychology, or unsupported causal claims. \\

\bottomrule
\end{tabular}
}
\caption{
Evaluation rubric for simulated user behaviors.
The rubric contains ten criteria grouped into four categories.
}
\label{tab:evaluation_rubric}
\end{table*}

The DPO judge receives the correct high-level action label but not the gold visible text.
This prevents the judge from rewarding exact surface matching and encourages evaluation of target grounding, action fidelity, persona consistency, memory discipline, stance, naturalness, specificity, format validity, and reasoning plausibility.
The weighted score is computed as
\[
S(\tau)
=
\frac{
2R_1+\sum_{i=2}^{10}R_i
}{11},
\]
where action fidelity is double-weighted because choosing the correct action family remains a central fidelity requirement after target selection.

\subsection{Rubric Activation Analysis}
\label{app:rubric_activation}

\begin{figure}[t]
\centering
\includegraphics[width=\columnwidth]{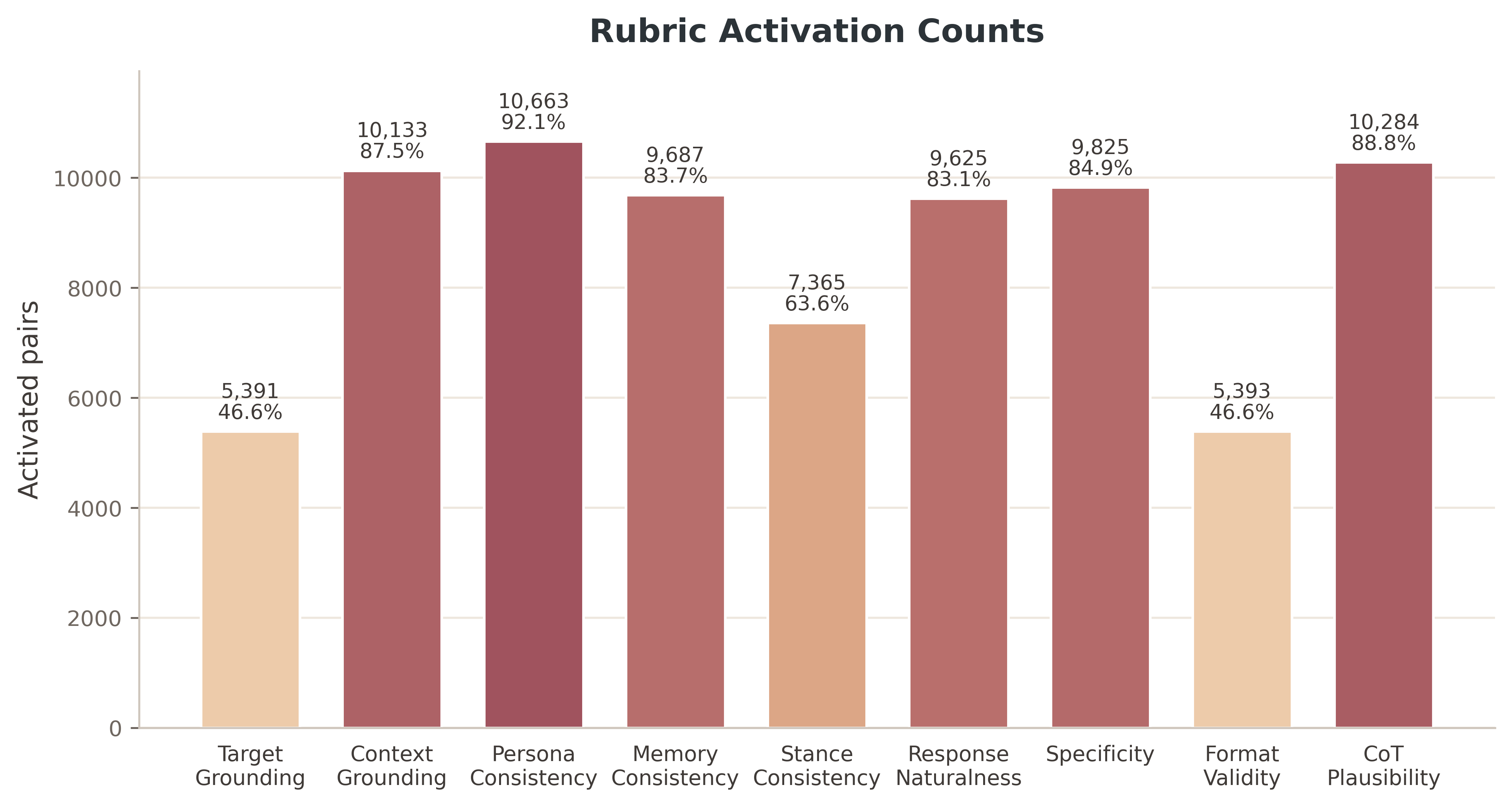}
\caption{
Activation counts of rubric dimensions over the evaluation set.
The numbers above each bar show the absolute number of activated pairs and their percentage in the full annotated set.
}
\label{fig:rubric_activation}
\end{figure}

Figure~\ref{fig:rubric_activation} reports how frequently each rubric dimension is activated in the annotated evaluation pairs.
Persona consistency, CoT plausibility, and context grounding are the most frequently activated dimensions, while response naturalness, specificity, and memory consistency are also highly active.
This shows that the evaluation is not merely action-based, but also depends on profile alignment, social-media realism, and reasoning plausibility.

\section{Additional Analyses}
\label{app:additional_analysis}

This section reports additional analyses that complement the main experimental results.
We include platform-level action statistics, thread-level structural properties, the effect of memory retrieval size, judge robustness, and pairwise group-simulation preference results.

\subsection{Platform-level Simulation Statistics}
\label{app:platform_statistics}

\begin{figure}[ht]
    \centering
    \includegraphics[width=0.62\columnwidth]{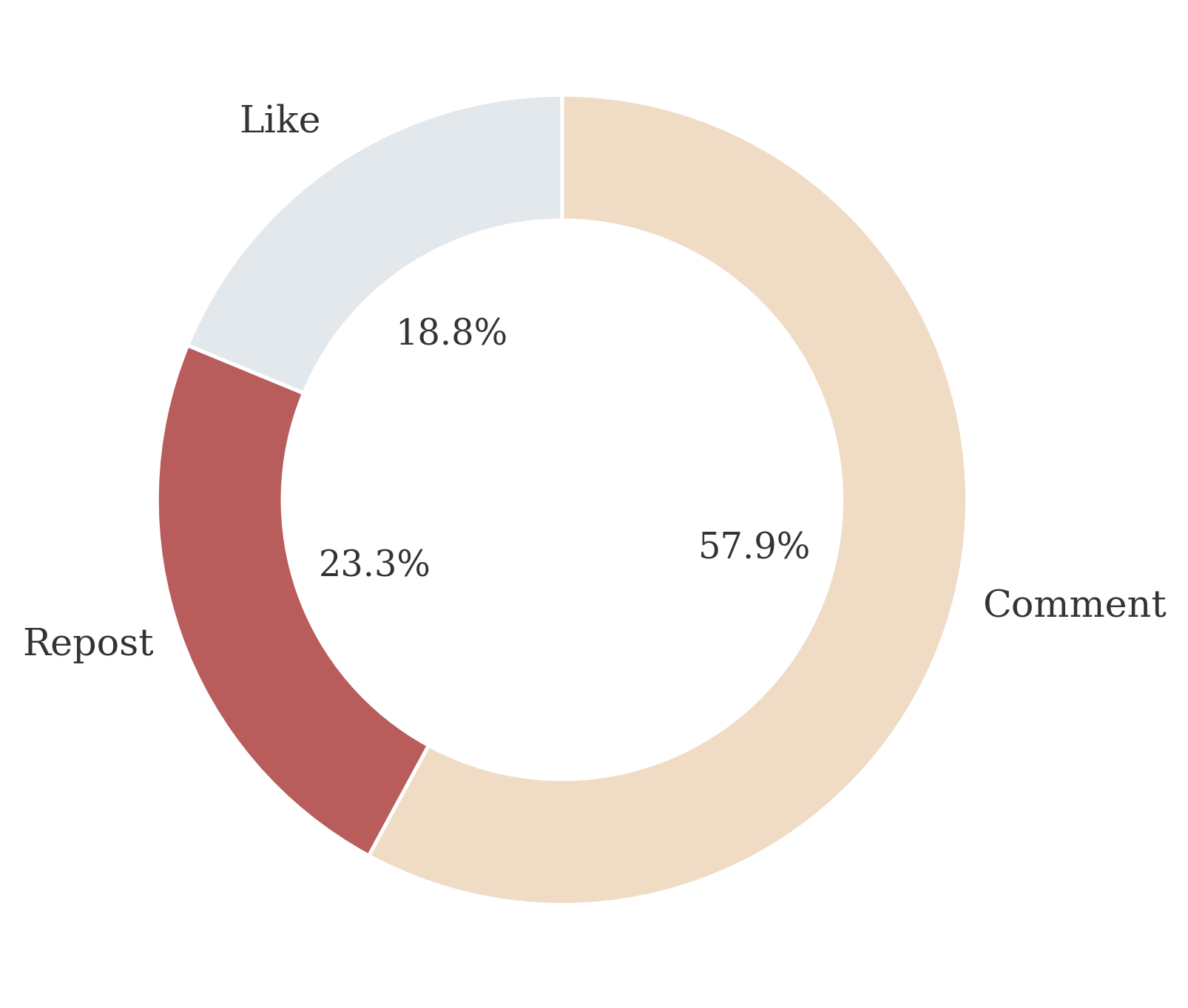}
    \caption{
    Distribution of simulated agent actions.
    }
    \label{fig:app_action_distribution}
\end{figure}

Figure~\ref{fig:app_action_distribution} shows the distribution of simulated agent actions.
Comments account for the largest proportion of actions, followed by reposts and likes.
This indicates that the simulated platform is primarily driven by active conversational behavior rather than passive engagement.

\begin{figure}[ht]
    \centering
    \includegraphics[width=\columnwidth]{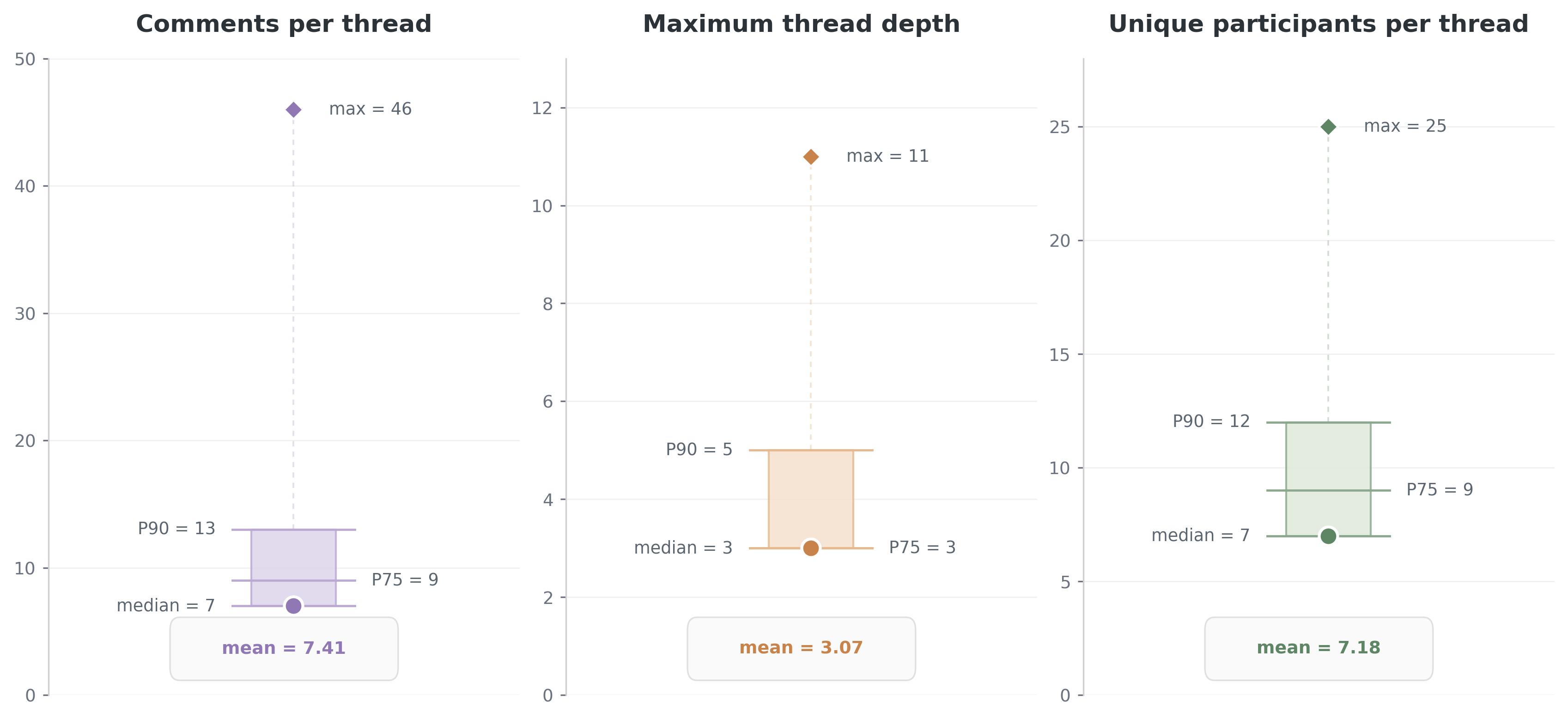}
    \caption{
    Statistical properties of simulated discussion threads.
    }
    \label{fig:app_thread_statistics}
\end{figure}

Figure~\ref{fig:app_thread_statistics} presents thread-level interaction statistics, including comments per thread, maximum thread depth, and unique participants.
Most discussions remain moderate in scale, while a small subset develops into deeper and more active interactions.
This suggests that the simulator does not simply generate uniformly dense discussion trees, but instead produces a mixture of shallow and deeper interaction patterns.

\subsection{Effect of Memory Retrieval Size}
\label{app:memory_retrieval_size}

\begin{figure}[ht]
    \centering
    \includegraphics[width=\columnwidth]{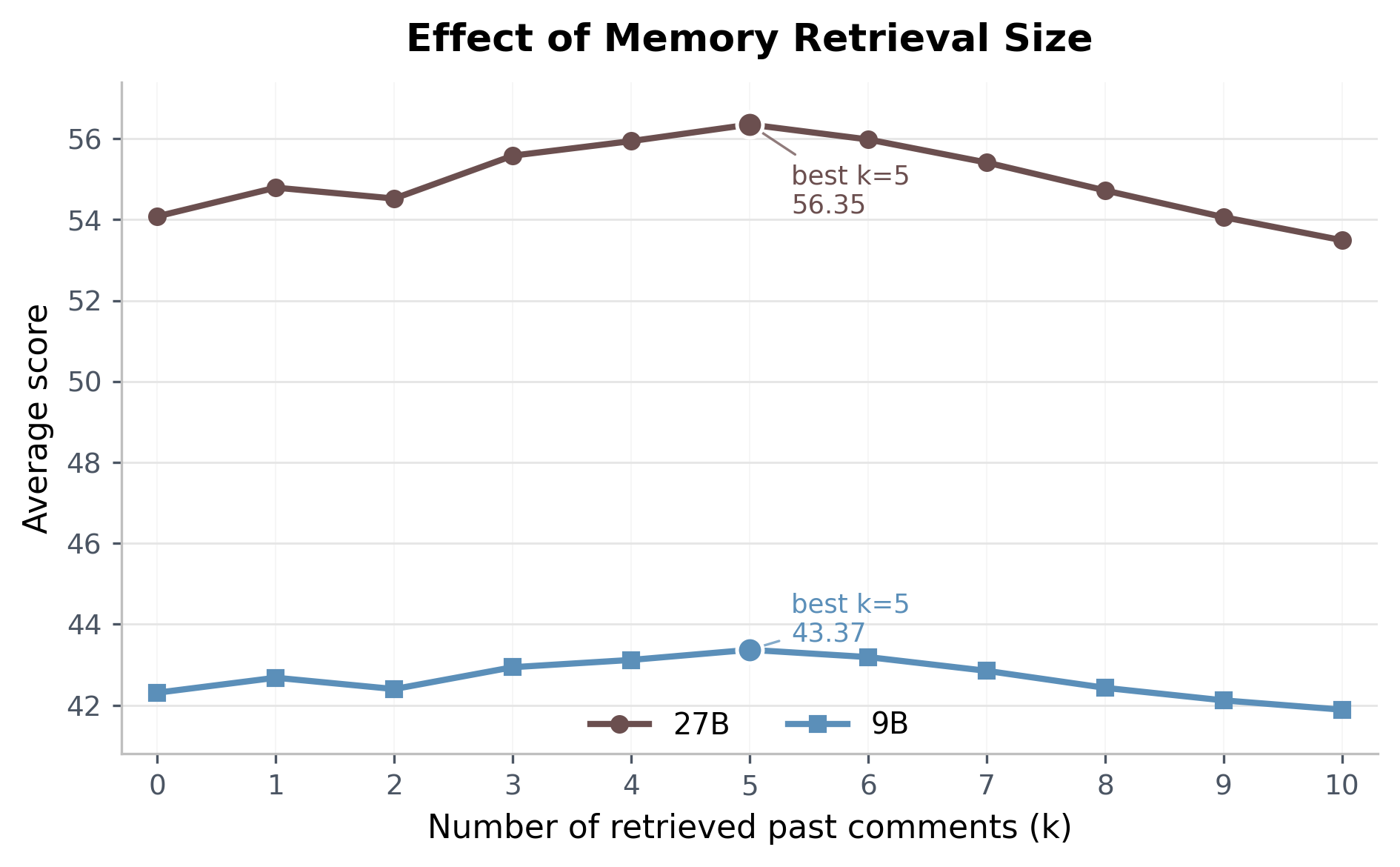}
    \caption{
    Effect of memory retrieval size $k$ on average generation quality.
    }
    \label{fig:memory_retrieval_size}
\end{figure}

We vary the number of retrieved past comments $k$ and evaluate the average generation quality.
As shown in Figure~\ref{fig:memory_retrieval_size}, both the 9B and 27B models achieve the best performance when $k=5$.
Smaller values provide insufficient historical context, while larger values may introduce redundant or noisy memories.
Therefore, we set $k=5$ as the default retrieval size.

\subsection{Judge Robustness}
\label{app:judge_self_preference}

LLM-as-a-judge evaluation may introduce evaluator-specific preferences.
To examine whether our conclusion depends on a particular evaluator, we evaluate the same anonymized model outputs using multiple judge models from different model families.
As shown in Table~\ref{tab:judge_self_preference}, PsyMemBehav 27B receives the highest score under all reported judges, suggesting that its advantage is not caused by a single evaluator or judge-family preference.

\begin{table}[ht]
\centering
\scriptsize
\setlength{\tabcolsep}{4pt}
\renewcommand{\arraystretch}{1.10}
\resizebox{\columnwidth}{!}{
\begin{tabular}{lcccc}
\toprule
\textbf{Judge}
& \textbf{PsyMemBehav}
& \textbf{GPT-5.4}
& \textbf{Gemini}
& \textbf{Claude} \\
\midrule
Qwen3.5-27B
& \textbf{61.96} & 54.23 & 52.02 & 47.60 \\
GPT-5.4
& \textbf{63.09} & 60.06 & 48.94 & 57.12 \\
Gemini-2.5-Pro
& \textbf{72.50} & 67.60 & 62.75 & 61.33 \\
Claude-Sonnet-4.6
& \textbf{47.27} & 40.68 & 33.88 & 44.44 \\
\bottomrule
\end{tabular}
}
\caption{
Judge robustness analysis.
Each row corresponds to one judge model; each column corresponds to an evaluated model.
PsyMemBehav 27B achieves the highest score under all four judges.
}
\label{tab:judge_self_preference}
\end{table}

We also conduct a direct pairwise group-simulation preference test between PsyMemBehav 27B and GPT-5.4.
For each instance, both models are given the same real thread seed and asked to generate a group-level continuation.
The better simulation is selected according to its fidelity to the real discussion dynamics.

\begin{table}[ht]
\centering
\scriptsize
\setlength{\tabcolsep}{8pt}
\renewcommand{\arraystretch}{1.12}
\begin{tabular}{lc}
\toprule
\textbf{Model} & \textbf{Win Rate} \\
\midrule
GPT-5.4 & 34.0\% \\
\rowcolor{oursblue}
\textbf{PsyMemBehav 27B} & \textbf{64.0\%} \\
Tie & 2.0\% \\
\bottomrule
\end{tabular}
\caption{
Pairwise group simulation preference test between GPT-5.4 and PsyMemBehav 27B.
}
\label{tab:pairwise_group_sim_winrate}
\end{table}

\subsection{Human--Gemini Evaluation Agreement}
\label{app:human_gemini_agreement}

For the human evaluation, we selected 30 anonymized user IDs and evaluated five behavior instances per ID, yielding 150 annotated instances. Human evaluators and Gemini scored the same outputs under the same seven-dimensional rubric. Table~2 in the main paper compares their dimension-level means. The human and Gemini overall means are 54.40 and 55.46, respectively, a human-minus-Gemini difference of $-1.06$ points. Humans score semantic alignment, style fidelity, and naturalness slightly higher, whereas Gemini scores intent, emotion, stance, and profile/memory consistency higher. This dimension-mean comparison assesses aggregate calibration; it neither establishes per-instance agreement nor replaces a formal human reliability analysis.

The largest absolute gaps occur for stance (4.20 points) and emotion (3.60 points), while all other dimensions differ by at most 3.73 points. Because the signs vary, Gemini does not show a simple global tendency to be uniformly more lenient or stringent. The dimension-specific gaps therefore motivate reporting human evaluation together with LLM-judge results and cross-judge robustness.

\subsection{Dataset Scale and Longitudinal Coverage}
\label{app:dataset_scale}

Our use of \emph{large-scale} does not imply platform-wide coverage. Instead, it refers to the joint scale of real-user coverage, total behavioral volume, and the density of longitudinal histories available for user modeling. We therefore characterize the resource more precisely as a \emph{large-scale, longitudinal, user-centered dataset}. Table~\ref{tab:dataset_scale_comparison} compares PsyMemBehav-bench with closely related work along three dimensions: the number of real users represented, the behavioral evidence available for modeling each user, and the granularity of the prediction or simulation target.

\begin{table*}[t]
\centering
\scriptsize
\setlength{\tabcolsep}{2.5pt}
\renewcommand{\arraystretch}{1.12}
\begin{tabular}{
>{\raggedright\arraybackslash}p{0.24\textwidth}
>{\raggedright\arraybackslash}p{0.16\textwidth}
>{\raggedright\arraybackslash}p{0.24\textwidth}
>{\raggedright\arraybackslash}p{0.27\textwidth}}
\toprule
\textbf{Study} & \textbf{Real-user coverage} & \textbf{Behavioral evidence} & \textbf{Prediction/simulation target} \\
\midrule
FineRob: \emph{LLM-Based Behavior Prediction for Social Media Users with Continuous Memory}~\citep{li2025finerob}
& 1,866 users
& 78.6K fine-grained prediction records; 126 historical records per user on average
& Interaction object, behavior type, and content \\

\emph{Can LLMs Simulate Social Media Engagement? A Study on Action-Guided Response Generation}~\citep{qiu2025engagement}
& User count not reported
& 3,990 instances; 1--31 historical examples per user
& Engagement-action selection and response generation \\

MOSAIC~\citep{liu2025mosaic}
& 204 real users and 204 simulated personas
& 30 posts plus survey-derived personas
& Group-level content dissemination and regulation \\

GA-S$^3$~\citep{zhang2025gas3}
& Group agents rather than identifiable real-user histories
& 30 events over seven days
& Aggregate traffic and collective behavior \\

\rowcolor{oursblue}
\textbf{PsyMemBehav-bench}
& \textbf{5,862 real users}
& \textbf{5.57M observed behaviors; approximately 950 behaviors per user}
& \textbf{User-level interaction target, target-conditioned action, and response} \\
\bottomrule
\end{tabular}
\caption{Scale comparison with closely related social-media behavior prediction and simulation studies.}
\label{tab:dataset_scale_comparison}
\end{table*}

Here, \emph{real-user coverage} denotes the number of distinct real users whose observed behavioral evidence is used to construct or evaluate the simulator. \emph{Behavioral evidence} captures both the total amount of data and the density of the history available for each user. \emph{Prediction/simulation target} distinguishes user-level target--action--response prediction from aggregate diffusion or traffic simulation.

Among the closely related studies that report real-user counts, PsyMemBehav-bench contains more real users and substantially denser longitudinal behavioral histories. Its scale characterization is therefore based not on the number of users alone, but on the combination of 5,862 real users, 5.57 million observed behaviors, and approximately 950 behaviors per user. At the same time, these statistics should not be interpreted as platform-wide coverage or as a representative sample of all social-media users. The collection procedure favors observable users with sufficiently rich histories and may therefore overrepresent highly active accounts while underrepresenting low-activity, private, or newly created accounts. Platform access constraints, the selected observation period, and activity-based sampling can further shape the resulting population. These limitations should be considered when interpreting generalization beyond the studied users and contexts.

\section{Evaluation Metric Definitions}
\label{app:metrics}

\subsection{Profile Fidelity: Normalized Edit Distance}
\label{app:ned}

Profile fidelity measures how faithfully a simulator reproduces a user's behavioral profile.
For each user, we re-annotate the simulator's generated behaviors using the same annotation model $g_{\mathrm{prof}}$ to obtain a predicted profile vector $\hat{P}_u$, and compare it against the original profile $P_u$ derived from ground-truth history.

We report fidelity using $100\times(1 - \mathrm{NED})$, where NED is the Normalized Edit Distance between the discretized profile sequences:
\[
\mathrm{NED}(P_u, \hat{P}_u)
= \frac{\mathrm{ED}(P_u, \hat{P}_u)}{\max(|P_u|, |\hat{P}_u|)},
\]
where $\mathrm{ED}(\cdot,\cdot)$ denotes the Levenshtein edit distance and $|\cdot|$ denotes sequence length after discretizing continuous trait scores into ordinal bins. Because both reference and generated behaviors are interpreted through the same annotator, this score measures relative cross-system consistency under a fixed profiling lens; it should not be interpreted as absolute psychometric validity.
A score of $100\times(1-\mathrm{NED}) = 100$ indicates a perfect match; higher values reflect closer alignment between simulated and real user profiles.
We report two profile-fidelity scores separately: one for general psychological traits ($P_u^{\mathrm{gen}}$) and one for platform-specific behavioral traits ($P_u^{\mathrm{plat}}$).

\subsection{Group-Level Simulation Metrics}
\label{app:group_metrics}

The group-level evaluation places multiple profiled simulators into real thread seeds and measures how closely the generated interaction structure matches real comment-section continuations.

\paragraph{Reply Reciprocity (Rec.).}
Reciprocity is a standard directed-network measure of mutual linkage \citep{garlaschelli2004reciprocity}. We compute the fraction of interacting user pairs $(u,v)$ for which both reply directions occur:
\[
\mathrm{Rec.} = \frac{|\{(u,v): u\to v \wedge v\to u\}|}{|\{(u,v): u\to v \vee v\to u\}|}.
\]
Overly high reciprocity indicates an unrealistic group-chat pattern rather than broadcast discussion.

\paragraph{Stance Deviation (Stance Dev.).}
Following the conventional favor/against/neither formulation of stance \citep{mohammad2016semeval}, each comment is annotated as \textsc{Pro}, \textsc{Anti}, or \textsc{Neutral}. We measure the $\ell_1$ distance between simulated and real stance distributions:
\[
\mathrm{StanceDev} = \sum_{s \in \{\text{Pro, Anti, Neutral}\}}
\bigl|p_{\mathrm{sim}}(s) - p_{\mathrm{real}}(s)\bigr|.
\]
Lower values indicate closer reproduction of collective opinion composition.

\paragraph{Attention Concentration (Top1 KOL / Top5 PR).}
Top1 KOL is the fraction of all received replies captured by the single most-addressed user, reflecting key-opinion-leader concentration \citep{katz1955personal}. Top5 PR is the sum of normalized PageRank scores \citep{brin1998anatomy} for the five highest-ranked users in the directed reply graph. Together they test whether simulated attention and structural influence concentrate as in the real thread.

\paragraph{Overall Deviation (MAE / MSE).}
MAE and MSE aggregate, respectively, absolute and squared per-thread deviations between simulated and real statistics. Lower values indicate closer overall group-level fidelity; MSE penalizes large mismatches more strongly.

\section{Case Studies}
\label{app:case_studies}

This section provides qualitative and group-level examples that complement the aggregate results in the main paper.
We include both a single-reply comparison and a replay-style group simulation comparison.
The single-reply case highlights how different models vary in stance commitment and response style under the same thread context.
The group-level case shows that a model can produce plausible individual replies while still failing to reproduce the collective dynamics of a real discussion.

\subsection{Single-reply Comparison}
\label{app:single_reply_case}

Table~\ref{tab:single_reply_case} compares different models under the same thread context and target comment.
The real user reply is cautious and low-commitment.
Several strong general-purpose models produce more socially reactive or stance-bearing responses, while PsyMemBehav 27B remains more hedged and balancing.

\begin{table}[ht]
\centering
\small
\setlength{\tabcolsep}{3pt}
\renewcommand{\arraystretch}{1.12}
\resizebox{\columnwidth}{!}{%
\begin{tabular}{
>{\raggedright\arraybackslash}p{2.0cm}
>{\raggedright\arraybackslash}p{5.8cm}
>{\centering\arraybackslash}p{1.8cm}
>{\raggedright\arraybackslash}p{3.6cm}
}
\toprule
\textbf{Model}
& \textbf{Generated Reply}
& \textbf{Stance}
& \textbf{Observation} \\
\midrule

Real user
& ``I can see why people read the timing as suspicious, but it may also just be production timing.''
& Neutral
& Conservative and low-commitment stance. \\

Gemini-3.1-Pro
& ``They could still be reworking it. This week gave them too much new material to ignore.''
& Neutral
& Cautious and relatively close to the real style. \\

GPT-5.4
& ``That timing feels awfully convenient. People are going to read into it no matter what.''
& Neutral / mildly charged
& More socially reactive and discussion-provoking. \\

Claude-Sonnet-4.6
& ``That explanation feels way too convenient given the pressure around this.''
& Pro / supportive
& More likely to clearly agree with a suspicious reading. \\

Qwen3.7-Max
& ``That excuse is transparent. They probably got spooked by legal threats or the network.''
& Pro / charged
& Stronger stance and more speculative. \\

\rowcolor{oursblue}
\textbf{PsyMemBehav 27B}
& ``The reality is probably more mixed than this thread is making it.''
& Neutral
& More hedged and balancing, closer to the real user's conservative stance. \\

DeepSeek-V4-Pro
& ``There is a narrower point here, but the broad claim feels overcooked.''
& Anti / neutral
& Often gives a cooling or corrective reply. \\

\bottomrule
\end{tabular}}
\caption{
Single-reply comparison under the same thread context.
}
\label{tab:single_reply_case}
\end{table}

\subsection{Group Simulation Comparison}
\label{app:group_sim_case}

The group simulation setting starts from the same real thread seed and asks each model to generate a continuation involving multiple agents.
Table~4 in the main paper reports aggregate interaction and network-level results over the evaluation set. Table~\ref{tab:group_sim_case} complements that aggregate evaluation with the stance distribution of this representative replay case. Each comment is assigned to one of three mutually exclusive stance classes: \textbf{Pro} explicitly supports the focal claim, \textbf{Anti} opposes it, and \textbf{Neutral (Neu.)} expresses neither clear support nor opposition; their proportions therefore sum to one. \textbf{Charged (Chg.)} is the share of stance-bearing comments, $\mathrm{Chg.}=p(\mathrm{Pro})+p(\mathrm{Anti})=1-p(\mathrm{Neutral})$. \textbf{Distance (Dist.)} is the $\ell_1$ distance between a model's stance vector and the real continuation, $\sum_s|p_{\mathrm{sim}}(s)-p_{\mathrm{real}}(s)|$; it ranges from 0 to 2, and lower is better. \textbf{Entropy (Ent.)} is normalized Shannon entropy, $-\sum_s p(s)\log p(s)/\log 3$; it ranges from 0 (all comments in one stance) to 1 (a uniform three-way distribution). Because entropy measures stance diversity rather than accuracy, neither a larger nor a smaller raw value is intrinsically better; closeness to the real entropy indicates better fidelity.

\begin{table}[!t]
\centering
\scriptsize
\setlength{\tabcolsep}{3pt}
\renewcommand{\arraystretch}{1.10}
\resizebox{\columnwidth}{!}{%
\begin{tabular}{lcccccc}
\toprule
\textbf{Model}
& \textbf{Pro}
& \textbf{Anti}
& \textbf{Neu.}
& \textbf{Chg.}
& \textbf{Dist.}
& \textbf{Ent.} \\
\midrule
Real
& 0.085 & 0.085 & 0.830 & 0.170 & 0.000 & 0.523 \\
Gemini-3.1-Pro
& 0.312 & 0.125 & 0.563 & 0.437 & 0.534 & 0.862 \\
GPT-5.4
& 0.479 & \textbf{0.083} & 0.438 & 0.562 & 0.788 & 0.838 \\
Claude-Sonnet-4.6
& 0.322 & 0.212 & 0.466 & 0.534 & 0.728 & 0.955 \\
Qwen3.7-Max
& 0.417 & 0.062 & 0.521 & 0.479 & 0.664 & 0.798 \\
\rowcolor{oursblue}
\textbf{PsyMemBehav 27B}
& \textbf{0.182} & 0.121 & \textbf{0.697} & \textbf{0.303} & \textbf{0.266} & \textbf{0.744} \\
DeepSeek-V4-Pro
& 0.423 & 0.135 & 0.442 & 0.558 & 0.776 & 0.906 \\
\bottomrule
\end{tabular}}
\caption{
Stance-distribution comparison for a representative continuation from the same real thread seed.
Neu.\ = Neutral; Chg.\ = Charged; Dist.\ = stance distance; Ent.\ = entropy.
Aggregate interaction and network-level results are reported in Table~4 in the main paper; metric definitions are provided in the \emph{Group-Level Simulation Metrics} subsection.
}
\label{tab:group_sim_case}
\end{table}

\paragraph{Discussion.}
This representative case shows that strong general-purpose models generate substantially more stance-bearing responses than the real continuation. The real thread is predominantly neutral (0.830), with a charged share of 0.170. Among the simulated systems, PsyMemBehav 27B has the highest neutral share (0.697), the lowest charged share (0.303), and the smallest stance distance (0.266), making its stance distribution the closest to the real continuation. Its entropy (0.744) is also the closest to the real value (0.523). This case-level analysis complements, rather than duplicates, the aggregate thread-level comparison in Table~4 in the main paper.


\section{Prompt Templates}
\label{app:prompt_templates}

This section lists the main prompt templates used in PsyMemBehav.
We include the prompts for CoT validation, action generation, reference CoT construction, pairwise DPO judging, and seven-dimension semantic evaluation.
All prompt tables are shown as double-column tables for readability.

\begin{table*}[p]
\centering
\setlength{\tabcolsep}{4pt}
\renewcommand{\arraystretch}{0.95}
\begin{tabular}{p{0.96\textwidth}}
\hline \hline
\rowcolor[gray]{0.9}
\multicolumn{1}{>{\centering\arraybackslash}p{0.96\textwidth}}{\textbf{A Reference Prompt for CoT Validation.}} \\
\hline
\begin{minipage}{0.96\textwidth}
\begin{lstlisting}[
basicstyle=\fontsize{6.7pt}{6.9pt}\selectfont\ttfamily,
breaklines=true,
breakatwhitespace=false,
columns=fullflexible,
keepspaces=true,
showstringspaces=false,
frame=none,
aboveskip=2pt,
belowskip=2pt,
linewidth=0.95\textwidth
]
You are a validator for generated hidden CoT in a real social media user memory activation dataset.

You will receive the system prompt, dialogue prefix, original assistant target, and a generated assistant value containing <think> plus the original target.

Your job is to decide whether the <think> is valid supervision for deciding when to use the scene memory pool.

=== REFERENCE CONTEXT ===
{reference_content_with_memories}

=== ORIGINAL ASSISTANT TARGET ===
{target_assistant_content}

=== GENERATED ASSISTANT VALUE TO VALIDATE ===
{generated_reasoning}

Check the following:

1. Structure:
- There must be exactly one <think> block before the assistant target.
- The <think> should contain exactly three natural paragraphs.
- It should not use explicit awareness/activation/decision tags, headings, bullets, or numbered step labels.
- The assistant target after </think> must exactly match the original assistant target.

2. Source grounding:
- Every fact in the <think> must be supported by the system profile, current scenario, dialogue prefix, Relevant Past Memories, Other Characters Relationships, or original assistant target.
- No fabricated scene details, motives, actions, or emotions.

3. Current-turn boundary:
- The <think> must explain only the current assistant turn.
- It must not use later dialogue or future plot knowledge unavailable to the character at this story point.
- If the system profile contains future events, the CoT should not use them unless the current character could know them.

4. Information leakage:
- The <think> must not leak information unavailable to the character at the current turn.
- The <think> must not reveal information that the visible assistant target withholds.
- The <think> may use the original assistant target to check stance alignment and response policy, but it must not quote, closely paraphrase, or explicitly pre-announce the target's distinctive surface wording, actions, or inner thoughts.
- Treat future-biographical details in the profile as unsafe unless the character could know or act from them at this story point.

5. Memory decision:
- Independently decide whether the scene memory pool is needed.
- Scene memory pool means only the listed Relevant Past Memories, not the character profile.
- If the current dialogue, current scenario, and profile are sufficient, memory should be marked not needed.
- If the target requires information absent from current context/profile but present in a relevant memory, memory should be marked needed.
- Check that the generated CoT names the correct memory ids when memory is needed.
- When memory is not needed, the CoT should explain why the current dialogue, current scenario, and/or profile are sufficient.
- If memory is needed, check that the generated CoT explains how each activated memory affects behavior in the current turn.
- If multiple memories are activated, check that the generated CoT compares their directions: aligned, opposed, or mixed, and explains the final net effect.

6. Profile vs memory:
- If the required information is in the profile, the CoT should not claim that a scene memory is needed.
- It may say profile knowledge is needed, but scene memory pool is not.

7. Answer leakage:
- The <think> must not quote or closely paraphrase the final assistant wording.
- It may describe response function, tone, and constraints, but not the final surface form.

8. Training usefulness:
- The <think> should clearly analyze the current information need, memory relevance, and response policy.
- It should not be a generic template or a repetition of the final answer.
\end{lstlisting}
\end{minipage}
\\
\hline \hline
\end{tabular}
\caption{A reference prompt for CoT validation.}
\label{tab:cot_validation_prompt}
\end{table*}

\clearpage
\begin{table*}[p]
\centering
\setlength{\tabcolsep}{4pt}
\renewcommand{\arraystretch}{0.92}
\begin{tabular}{p{0.96\textwidth}}
\hline \hline
\rowcolor[gray]{0.9}
\multicolumn{1}{>{\centering\arraybackslash}p{0.96\textwidth}}{\textbf{Unified End-to-End Behavior-Prediction Prompt.}} \\
\hline
\begin{minipage}{0.96\textwidth}
\begin{lstlisting}[
basicstyle=\fontsize{6.4pt}{6.8pt}\selectfont\ttfamily,
breaklines=true,
breakatwhitespace=false,
columns=fullflexible,
keepspaces=true,
showstringspaces=false,
frame=none,
aboveskip=2pt,
belowskip=2pt,
linewidth=0.95\textwidth
]
You are roleplaying as social-media user {username}. Given the user profile, retrieved pre-event history, current thread, and candidate targets, first select the interaction target. Conditioned on that target, predict the user's recorded engagement as exactly one of Comment, Like, or Repost. Generate textual content only for Comment. Base both decisions on evidence in the input and do not sound like an assistant.

[Persona]
{rendered_profile}

[Retrieved Past Comments]
{rendered_history_block}

[Thread]
[post] @{post_author} -- "{post_title}"
{post_text}

[Comments]
- [{cand_id_1}] @{author_1} (reply to [post]): "{full_text_1}"
- [{cand_id_2}] @{author_2} (reply to [{parent_id}]): "{full_text_2}"

[Candidate Targets]
[post], [{cand_id_1}], [{cand_id_2}]

Rules:
- Select a valid target before choosing the action; neither the gold target nor the gold action is provided.
- Conditioned on the selected target, infer exactly one action from {Comment, Like, Repost} and keep the target fixed.
- Emit valid JSON for exactly one action and omit all fields belonging to the other actions.
- Comment: output action="Comment" and a reply object containing reply_to_id and natural, thread-specific text.
- Like: output action="LikeThread" and a target object containing id and kind; LikeThread and LikeBasic are evaluated as high-level Like. A like represents light endorsement, acknowledgment, or agreement.
- Repost: output action="Repost" and the same target object; a repost represents stronger public amplification than a like.
- Do not generate visible text for Like or Repost. Use history only as preference evidence and keep private reasoning concise and grounded.

Output exactly:
<think>
- Selected target: ...
- Predicted action/intensity for this target: Comment, Like, or Repost / ...
- Context grounding: ...
- Persona/memory/stance cue: ...
- Naturalness cue: ...
</think>

Then output exactly one valid <answer> branch; do not output branch labels.
Comment: <answer>{"activity_type":"BehaviorPrediction","action":"Comment","reply":{"reply_to_id":"<candidate id>","text":"..."}}</answer>
Like: <answer>{"activity_type":"BehaviorPrediction","action":"LikeThread","target":{"id":"<candidate id>","kind":"post | comment"}}</answer>
Repost: <answer>{"activity_type":"BehaviorPrediction","action":"Repost","target":{"id":"<candidate id>","kind":"post | comment"}}</answer>
\end{lstlisting}
\end{minipage}
\\
\hline \hline
\end{tabular}
\caption{Unified end-to-end behavior-prediction template. All inputs and target-first reasoning steps are shared; only the selected action and its required output fields differ.}
\label{tab:unified_behavior_prediction_prompt}
\end{table*}

\clearpage
\begin{table*}[p]
\centering
\setlength{\tabcolsep}{4pt}
\renewcommand{\arraystretch}{0.95}
\begin{tabular}{p{0.96\textwidth}}
\hline \hline
\rowcolor[gray]{0.9}
\multicolumn{1}{>{\centering\arraybackslash}p{0.96\textwidth}}{\textbf{A Reference Prompt for Reference CoT Construction.}} \\
\hline
\begin{minipage}{0.96\textwidth}
\begin{lstlisting}[
basicstyle=\fontsize{6.7pt}{7.2pt}\selectfont\ttfamily,
breaklines=true,
breakatwhitespace=false,
columns=fullflexible,
keepspaces=true,
showstringspaces=false,
frame=none,
aboveskip=2pt,
belowskip=2pt,
linewidth=0.95\textwidth
]
# Action-DPO Reference CoT Prompt

You are writing inverse reasoning for a persona-simulation training example.

You are given:
- the user's profile and retrieved history,
- the current post/thread context,
- the ground-truth action and answer block.

Your job is not to predict a new action. Your job is to explain, briefly and plausibly, why this specific user would take the provided ground-truth action in this situation.

## What To Cover

Write a compact natural-language <think> block. It should cover:
- Action fit: why this action type is natural for the user here.
- Interaction intensity: why the amount of engagement is calibrated.
- Context grounding: what in the current post/thread creates the opening.
- Persona and memory: which profile/history signals matter.
- Stance and tone: the concrete stance or affective register implied by the action.
- Format/target: keep the selected target fixed and include visible target references only when the chosen action format requires them.

## Constraints

- Do not invent private motives, fake memories, or unsupported ideology.
- Do not claim that the exact hidden ground-truth text is the only possible wording.
- For like/repost actions, do not over-explain as if the user wrote a comment.
- Keep the reasoning concise; this should be a training CoT, not an essay.
- Output only one <think>...</think> block. Do not output an <answer> block.

## Output Format

<think>
Action fit: ...
Interaction intensity: ...
Context grounding: ...
Persona and memory: ...
Stance and tone: ...
Format/target: ...
</think>
\end{lstlisting}
\end{minipage}
\\
\hline \hline
\end{tabular}
\caption{A reference prompt for constructing inverse CoT supervision from a known ground-truth action.}
\label{tab:reference_cot_prompt}
\end{table*}

\clearpage
\begin{table*}[p]
\centering
\setlength{\tabcolsep}{4pt}
\renewcommand{\arraystretch}{0.92}
\begin{tabular}{p{0.96\textwidth}}
\hline \hline
\rowcolor[gray]{0.9}
\multicolumn{1}{>{\centering\arraybackslash}p{0.96\textwidth}}{\textbf{A Reference Prompt for Pairwise DPO Judging.}} \\
\hline
\begin{minipage}{0.96\textwidth}
\begin{lstlisting}[
basicstyle=\fontsize{6.7pt}{7.2pt}\selectfont\ttfamily,
breaklines=true,
breakatwhitespace=false,
columns=fullflexible,
keepspaces=true,
showstringspaces=false,
frame=none,
aboveskip=1pt,
belowskip=1pt,
linewidth=0.95\textwidth
]
You are a strict pairwise preference judge for persona-simulation DPO data.

You will receive:
- the target user's profile and retrieved past comments;
- the current social-media thread and candidate targets;
- the correct high-level action type only: comment/reply, like, or repost;
- Candidate A and Candidate B, each containing CoT, target, action, and optional visible reply text.

Important:
- The ground-truth visible content is hidden. Do not assume exact wording.
- LikeThread and LikeBasic are both treated as high-level like.
- Judge target grounding first, then the action conditioned on that target, persona fit, memory use, stance, naturalness, specificity, format, and CoT plausibility.
- For comment/reply, judge whether the visible text is plausible for this user and context.
- For like/repost, penalize unnecessary visible comments.
- Do not reward length or polished assistant-like writing by itself.

Score each candidate from 1 to 5:
R1. Action Fidelity
R2. Interaction Intensity Calibration
R3. Target and Thread Grounding
R4. Persona Consistency
R5. Memory Activation Discipline
R6. Stance Consistency
R7. Response Naturalness
R8. Specificity
R9. Format Validity
R10. CoT Plausibility

Use the weighted score:
weighted_score = (2*R1 + R2 + R3 + R4 + R5 + R6 + R7 + R8 + R9 + R10) / 11

Choose A, B, tie, or drop.
Drop weak comparisons if:
- confidence < 4;
- score gap is too small;
- candidates are near-duplicates;
- the winner wins mainly by being longer;
- either candidate has invalid target, action, format, or malformed JSON;
- the CoT fabricates private psychology or forced memory.

Return JSON only:
{
  "sample_id": "...",
  "correct_action_type": "comment|like|repost|other",
  "candidate_a_scores": {
    "R1": 1, "R2": 1, "R3": 1, "R4": 1, "R5": 1,
    "R6": 1, "R7": 1, "R8": 1, "R9": 1, "R10": 1,
    "weighted_score": 0.0
  },
  "candidate_b_scores": {
    "R1": 1, "R2": 1, "R3": 1, "R4": 1, "R5": 1,
    "R6": 1, "R7": 1, "R8": 1, "R9": 1, "R10": 1,
    "weighted_score": 0.0
  },
  "chosen": "A|B|tie|drop",
  "confidence": 1,
  "score_gap": 0.0,
  "winner_reason": "brief evidence-bound reason",
  "loser_weakness": "main concrete weakness",
  "keep_for_dpo": true
}
\end{lstlisting}
\end{minipage}
\\
\hline \hline
\end{tabular}
\caption{A reference prompt for pairwise DPO preference judging.}
\label{tab:pairwise_dpo_judge_prompt}
\end{table*}

\clearpage
\begin{table*}[p]
\centering
\setlength{\tabcolsep}{4pt}
\renewcommand{\arraystretch}{0.92}
\begin{tabular}{p{0.96\textwidth}}
\hline \hline
\rowcolor[gray]{0.9}
\multicolumn{1}{>{\centering\arraybackslash}p{0.96\textwidth}}{\textbf{A Reference Prompt for Seven-Dimension Semantic Evaluation.}} \\
\hline
\begin{minipage}{0.96\textwidth}
\begin{lstlisting}[
basicstyle=\fontsize{6.15pt}{6.2pt}\selectfont\ttfamily,
breaklines=true,
breakatwhitespace=false,
columns=fullflexible,
keepspaces=true,
showstringspaces=false,
frame=none,
aboveskip=1pt,
belowskip=1pt,
linewidth=0.95\textwidth
]
You are a strict, consistent, and objective evaluator for social media user simulation.
Your task is to assess how well the model-generated comment simulates the target user's real response in the same social media context.
You will be given:
1. The original social media post
2. The target user's history, profile, or memory information
3. The target user's real comment, used as the ground truth
4. The model-generated comment
Important evaluation principles:
- The ground-truth comment is not the only possible correct response, but it represents the user's actual reaction in this context.
- Do not judge only by lexical overlap. Evaluate semantic meaning, interaction intent, emotion, stance, style, persona/memory consistency, and naturalness.
- If the generated comment uses different wording but expresses a similar meaning, attitude, emotion, and interaction goal, it can receive a high score.
- If the generated comment is semantically relevant but does not match the user's style, stance, or interaction intent, reduce the corresponding scores.
- If the generated comment fabricates the user's personal history, identity, preferences, relationships, or experiences, significantly reduce the persona_memory_consistency score.
- If the generated comment sounds like AI writing, is overly formal, overly explanatory, or resembles a summary, reduce naturalness and style_fidelity scores.
- Do not give a high score simply because the generated comment is more complete or grammatically fluent. The goal is whether it sounds like the target user.
- Be strict. A score above 90 should only be given when the generated comment is extremely close to the target user's real response.
- Avoid giving all dimensions similar scores. Evaluate each dimension independently.
[Input]
Original social media post: {source_post}
Target user history/profile/memory: {user_profile_or_history}
Ground-truth comment: {ground_truth_comment}
Model-generated comment: {generated_comment}
[Scoring Scale] Assign an integer score from 0 to 100 per dimension.
90-100: Nearly perfect match.
75-89: Strong match, only minor differences.
60-74: Moderate match, noticeable differences.
40-59: Weak match, substantial deviation.
20-39: Poor match.
0-19: Very poor; irrelevant, contradictory, or fabricated.
[Evaluation Dimensions]
1. semantic_alignment
Assess whether the generated comment expresses the same core meaning, topic, and information points as the ground-truth comment.
2. intent_alignment
Assess whether the generated comment fulfills the same social interaction purpose (e.g., supporting, opposing, joking, asking, answering, expressing emotion, sharing experience, giving advice).
3. emotion_consistency
Assess whether the generated comment matches the ground-truth in emotion type, direction, and intensity.
4. stance_consistency
Assess whether the generated comment expresses the same stance (supportive, opposed, skeptical, neutral, ironic, etc.) toward the post, event, or opinion.
5. style_fidelity
Assess whether the generated comment resembles the target user's language style: length, tone, punctuation, emoji usage, slang, sentence patterns, formality, and humor style.
6. persona_memory_consistency
Assess whether the generated comment is consistent with the target user's historical preferences, interests, values, and long-term behavioral tendencies, and avoids fabricating user facts.
7. naturalness
Assess whether the generated comment sounds like a real social media comment rather than AI-generated text, a summary, or an essay-like response.
[Overall Score]
overall_score = unweighted average of the seven dimension scores, rounded to two decimal places.
[Output] JSON only. No markdown. No explanatory preface. All dimension scores must be integers from 0 to 100.
{
  "semantic_alignment":         {"score": 0, "reason": ""},
  "intent_alignment":           {"score": 0, "reason": ""},
  "emotion_consistency":        {"score": 0, "reason": ""},
  "stance_consistency":         {"score": 0, "reason": ""},
  "style_fidelity":             {"score": 0, "reason": ""},
  "persona_memory_consistency": {"score": 0, "reason": ""},
  "naturalness":                {"score": 0, "reason": ""},
  "overall_score": 0.00,
  "summary": ""
}
\end{lstlisting}
\end{minipage}
\\
\hline \hline
\end{tabular}
\caption{A reference prompt for seven-dimension semantic evaluation.}
\label{tab:seven_dim_eval_prompt}
\end{table*}

\end{document}